\documentclass[final,1p,times]{elsarticle}

\usepackage{amssymb}
\usepackage{amsmath}

\usepackage{mathtools}
\usepackage{amsthm}
\usepackage{algorithmic}
\usepackage{algorithm}
\usepackage{enumitem}
\usepackage{overpic}
\usepackage{float}
\usepackage{hyperref}

\usepackage[capitalize]{cleveref}

\theoremstyle{plain}
\newtheorem{theorem}{Theorem}[section]
\newtheorem{proposition}[theorem]{Proposition}

\theoremstyle{definition}

\newtheorem{assumption}[theorem]{Assumption}
\theoremstyle{remark}
\newtheorem{remark}[theorem]{Remark}
\usepackage{booktabs}

\newcommand{\R}{\mathbb{R}}
\newcommand{\U}{\mathcal{U}}
\newcommand{\A}{\mathcal{A}}
\renewcommand{\P}{\mathcal{P}}
\newcommand{\G}{\mathcal{G}}
\newcommand{\E}{\mathbb{E}}
\usepackage{comment}
\usepackage{xcolor}
\newcommand{\pluseq}{\mathrel{+}=}
\def\thick{0.8}
\usepackage{wrapfig}

\journal{CMAME}

\begin{document}

\begin{frontmatter}



  \title{Multi-Level Monte Carlo Training of Neural Operators}

  \author[label1]{James Rowbottom\fnref{fn1}}
  \author[label2]{Stefania Fresca\fnref{fn1}}
  \author[label3]{Pietro Lio}
  \author[label1]{Carola-Bibiane Schönlieb}
  \author[label4]{Nicolas Boull\'e}

  \fntext[fn1]{Equal contribution}

  \affiliation[label1]{organization={Department of Applied Mathematics and Theoretical Physics, University of Cambridge},
    city={Cambridge},
    postcode={CB3 0WA},
    country={UK}}

  \affiliation[label2]{organization={Department of Mechanical Engineering, University of Washington},
    city={Seattle},
    postcode={WA 98195},
    country={USA}}

  \affiliation[label3]{organization={Department of Computer Science and Technology, University of Cambridge},
    city={Cambridge},
    postcode={CB3 0WA},
    country={UK}}

  \affiliation[label4]{organization={Department of Mathematics,
        Imperial College London},
    city={London},
    postcode={SW7 2AZ},
    country={UK}}

  \begin{abstract}
    Operator learning is a rapidly growing field that aims to approximate nonlinear operators related to partial differential equations (PDEs) using neural operators. These rely on discretization of input and output functions and are, usually, expensive to train for large-scale problems at high-resolution. Motivated by this, we present a Multi-Level Monte Carlo (MLMC) approach to train neural operators by leveraging a hierarchy of resolutions of function discretization. Our framework relies on using gradient corrections from fewer samples of fine-resolution data to decrease the computational cost of training while maintaining a high level accuracy. The proposed MLMC training procedure can be applied to any architecture accepting multi-resolution data. Our numerical experiments on a range of state-of-the-art models and test-cases demonstrate improved computational efficiency compared to traditional single-resolution training approaches, and highlight the existence of a Pareto curve between accuracy and computational time, related to the number of samples per resolution.
  \end{abstract}




\end{frontmatter}

\section{Introduction}
Neural operators have been recently introduced as a mathematical tool to learn infinite-dimensional operators, mapping functions to functions, and have been applied in many scientific areas to approximate the solution operator of partial differential equations (PDEs) or solve inverse problems~\cite{li2021fourier,lu2021learning,kovachki2023neural,boulle2024mathematical,KOVACHKI2024419}. Several neural operator architectures have been proposed  including Fourier neural operators (FNOs)~\cite{li2021fourier}, deep operator networks (DeepONets)~\cite{lu2021learning}, graph neural operators~\cite{pfaff2020learning,anandkumar2020neural} and transformers~\cite{cao2021choose,hao2023gnot}. These networks exploit (or enforce) different properties of operators such as regularity of solutions and geometry of the underlying spatial domain on which functions are defined, and are supported by various approximation theory results~\cite{kovachki2021universal,KOVACHKI2024419,lanthaler2022error}.

The simulation of physical systems often relies on mesh-based representations, where the choice of mesh resolution presents a fundamental trade-off between computational cost and accuracy (this appears explicitly in, e.g., finite element error estimations~\cite{brenner2008mathematical}). Approximated solutions discretized on fine meshes are more accurate, but they significantly increase computational overhead during both training and inference. Although traditional deep learning approaches typically operate at a fixed resolution, limiting their ability to efficiently learn across multiple scales, neural operators are generally designed to process functions defined on arbitrary meshes by learning a representation with respect to a given basis~\cite{li2021fourier,bouziani2024structure} and interpolating/projecting across different resolutions. One of the main challenges in scaling neural operators to large-scale three-dimensional problems is the computational cost of the training phase, due to the fine gradient estimations needed to update the parameters of the neural network with a gradient-based optimization algorithm. Little work has been done in the literature on efficient training of neural operators, with the exception of a few works that proposed multi-scale architectures~\cite{li2020multipole,hemgno,bouziani2024structure} or adaptive subsampling~\cite{lanthaler2024discretization}.

\begin{figure}[t]
  \centering
  \begin{overpic}[width=0.8\textwidth, clip]{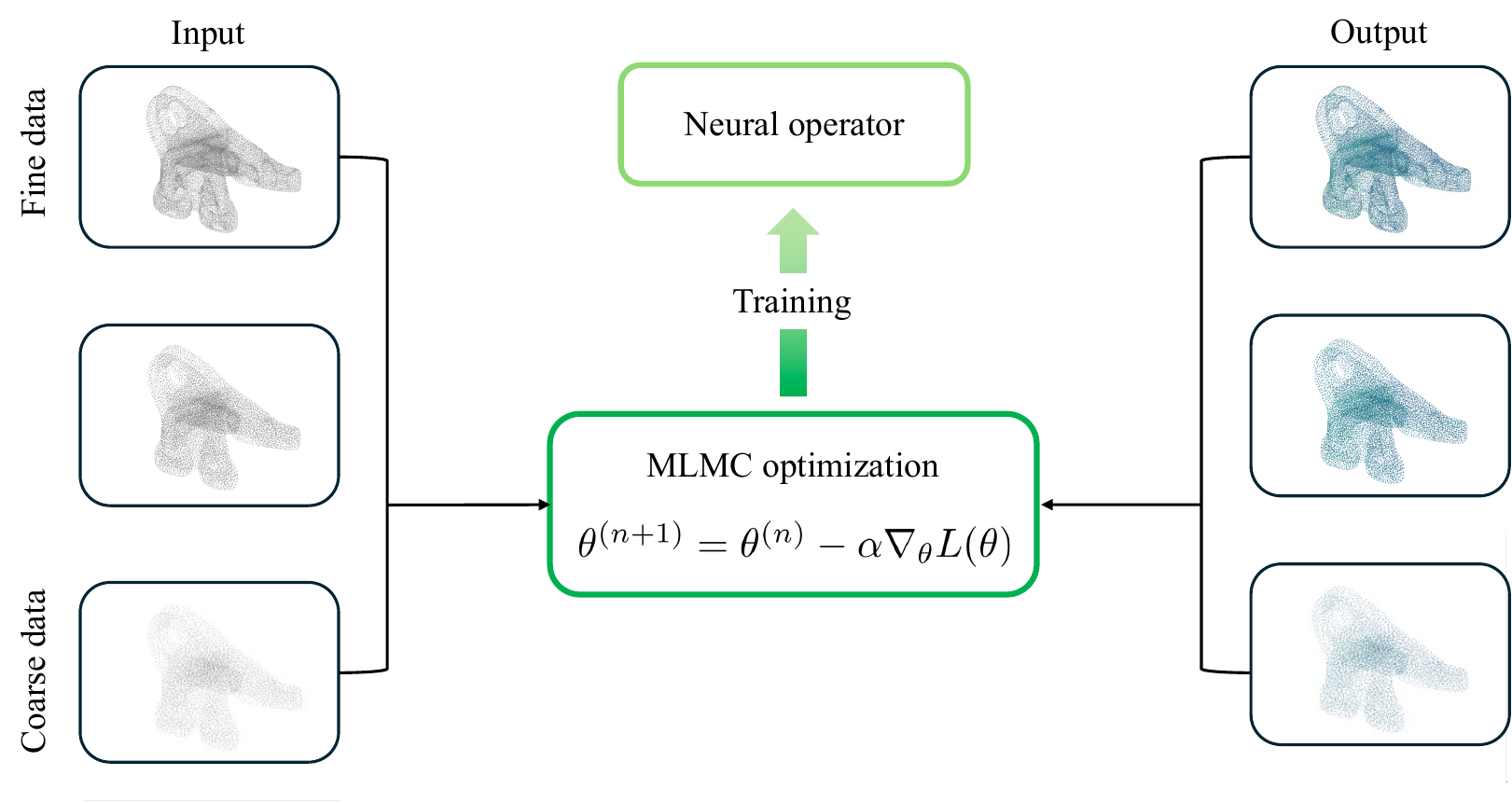}
  \end{overpic}
  \caption{Schematic diagram of the Multi-Level Monte Carlo (MLMC) approach for training neural operators. The input and output training data is downsampled from fine to coarse resolutions. The MLMC gradient estimation in \cref{eq_MLMC} combines samples from different levels to reduce the overall training cost while maintaining accuracy.}
  \label{fig:diagram}
\end{figure}

This work addresses the problem of the high training cost of neural operators on high-resolution data by proposing a flexible gradient estimation algorithm based on Multi-Level Monte Carlo (MLMC) methods~\cite{giles2008multilevel,giles2009multilevel,cliffe_multilevel_2011,giles2015multilevel} cast in the framework of mini-batch stochastic optimization. Our framework allows us to estimate the expectation of the gradients of the loss function on the finest resolution with respect to the neural network parameters using a multi-resolution approach. This quantity is expressed as a telescopic sum of gradients expectations across mesh levels. The aim is to drastically decrease the training computational time by capturing most of the dynamics, i.e., the variance, on the coarse grid, hence decreasing the number of gradient computations needed on the finest grid. We proved that the main assumptions of the MLMC theory are satisfied, under suitable regularity properties, by neural operators thus providing a robust mathematical justification in its application to operator learning. Moreover, our algorithm introduces a notion of ``Pareto optimality'' in the training process, where one can adjust the number of resolution levels (and hence the number of samples across levels) to achieve a trade-off between training time computational cost and accuracy.

Our work translates the classic MLMC methods for PDEs~\cite{giles2008multilevel,giles2009multilevel,cliffe_multilevel_2011,giles2015multilevel} to the context of neural operators, where the regularity properties of the PDE solution theoretically justify the approximation of the gradients. MLMC methods also inspired the development of several related works in machine learning. In \cite{li_multi-level_2021} an active learning strategy is employed to dynamically query an acquisition function to process new examples at different resolutions. \cite{lye_multi-level_2020} proposes a multi-level approach to learn input parameters to observable maps and apply it to uncertainty quantification. Moreover, \cite{jin_minimax_2023} studies a multi-level operator learning algorithm to obtain theoretical sample complexity estimates. More recently, \cite{gal2025towards,zakariaei2025multiscale} introduce a multi-scale stochastic gradient descent approach to train convolutional and graph neural networks at different resolutions. In particular, \cite{zakariaei2025multiscale} focuses on multi-scale training for standard CNNs applied to image processing tasks, while our work targets neural operators for challenging PDE problems using a variety of neural operator architectures.

While MLMC is a classical variance reduction technique, its extension to neural operator training is challenging. We develop a rigorous mathematical framework in \cref{sec:math_fram} that adapts MLMC theory to the operator learning setting, proving that neural operators satisfy key MLMC assumptions (cf.~\cref{prop:MLMC}). Building on this theory, we embed the MLMC estimator into mini-batch stochastic optimization by devising novel multi-resolution sampling and batching strategies in \cref{sec_mini_batch}. This includes introducing sample allocation formulas and sub-sampling schemes, which together enable efficient and consistent multi-resolution gradient estimation within each mini-batch. Implementing these ideas required significant engineering effort; indeed, we designed a cache-aware, memory-efficient multi-resolution batcher that manages data across mesh levels and minimizes costly memory transfers. Finally, we demonstrate the generality and effectiveness of our approach through extensive benchmarks in \cref{sec_experiments} on diverse state-of-the-art neural operator architectures (FNO, a message-passing PDE solver, and GINOT) and multiple challenging datasets (including 2D Darcy flow, time-dependent Navier–Stokes, and a 3D jet-engine bracket problem). These experiments show that our MLMC-based training scheme consistently accelerates convergence with minimal loss in accuracy.

\section{Multi-Level Monte Carlo method} \label{sec_method}
Our approach extends the MLMC framework to neural networks training. In particular, we aim at accelerating the training of neural operators by using the MLMC paradigm. MLMC methods stand on the idea that most of the uncertainty can be captured on the coarse grids and so the number of samples needed on the finest grid is greatly reduced.

\subsection{Mathematical Framework}
\label{sec:math_fram}
We follow the neural operator framework introduced by~\cite{kovachki2023neural} and consider an input function space $\mathcal{A} = L^2(\Omega_A)$ and an output Hilbert space of functions $\U$, respectively defined on bounded domains $\Omega_{\A}\subset\R^{d}$ and $\Omega_{\U}\subset\R^{d}$. Here, $\U$ is typically a (high-order) Sobolev space $H^k(\Omega_\U)$ for some $k\geq 0$. We aim to approximate a (nonlinear) operator $\G^\dagger:\A\to\U$ mapping between these spaces using a neural operator $\G_\theta:\A\to\U$, parametrized by a finite number of parameters $\theta\in\R^p$. We train the neural operator using pairs of observations $\{a^{(j)},u^{(j)}\}_{j=1}^N$, where $a^{(j)}\sim \mu$ are sampled i.i.d. from a probability measure $\mu$ supported on $\A$, and compactly supported on $H^{k}(\Omega_\A)$ for some $k\geq 0$. We aim to control the $L^2_\mu(\A;\U)$ Bochner norm as
\[
  \|\G^\dagger - \G_\theta\|_{L^2_\mu(\A;\U)}^2 = \E_{a\sim \mu}\|\G^\dagger(a) - \G_\theta(a)\|_{\U}^2=\int_\A \|\G^\dagger(a) - \G_\theta(a)\|_{\U}^2\,\textup{d}\mu(a).
\]
In this section, we assume that the operators $\G^\dagger$ and $\G_\theta$ satisfies the regularity conditions given by \cref{ass_reg_G}. While these are strong assumptions, they are standard in the operator learning approximation theory literature~\cite{kovachki2021universal,bouziani2024structure} and are necessary to control the image of the source term distribution through the nonlinear operator $\G^\dagger$.

\begin{assumption}[Regularity of $\G^\dagger$ and $\G_\theta$] \label{ass_reg_G}
  The operators $\G^\dagger$ and $\G_\theta$ are $\mu$-measurable, and belong to $L^2_\mu(\A;\U)$. Moreover, $\nabla_{\theta_i} \G_\theta\in L^2_\mu(\A;\U)$, for all $1\leq i\leq p$. We assume that the operators $\G^\dagger$, $\G_\theta$ are Lipschitz continuous from $\A\to\mathcal{U}$. Moreover, we assume that each component of $\nabla_\theta\G_\theta$ (with respect to $1\leq s\leq p$) is Lipschitz continuous from $\A\to\mathcal{U}$, with constant $L_\theta$.
\end{assumption}

We wish to solve the following optimization problem:
\[
  \min_{\theta\in\R^p}\E_{a\sim \mu}\|\G^\dagger(a) - \G_\theta(a)\|_{\U}^2,
\]
using a gradient-based algorithm as
\begin{equation} \label{eq:SGD}
  \begin{aligned}
    \theta^{k+1} & = \theta^k - \alpha \nabla_\theta (\E_{a\sim \mu}[\|\G^\dagger(a) - \G_\theta(a)\|_{\U}^2]) \\
                 & = \theta^k - \alpha \E_{a\sim \mu}[\nabla_\theta \|\G^\dagger(a) - \G_\theta(a)\|_{\U}^2],
  \end{aligned}
\end{equation}
where the second equality comes from the dominated convergence theorem. Hence, under \cref{ass_reg_G}, $\nabla_{\theta_i} \G_\theta\in L^2_\mu(\A;\U)$, for all $1\leq i\leq p$, and if $a\sim\mu$, Cauchy--Schwarz inequality yields
\[
  |\nabla_{\theta_i} \|\G^\dagger(a) - \G_\theta(a)\|_{\U}^2|  = 2|\langle \G^\dagger(a) - \G_\theta(a), \nabla_{\theta_i} \G_\theta(a)\rangle_{\U}| \leq 2\|\G^\dagger(a) - \G_\theta(a)\|_{\U}\|\nabla_{\theta_i} \G_\theta(a)\|_{\U}.
\]
Following Cauchy--Schwarz inequality on the expectation of the product of random variables, we can bound the expectation of the gradient as
\[
  \E_{a\sim \mu}[|\nabla_{\theta_i} \|\G^\dagger(a) - \G_\theta(a)\|_{\U}^2|] \leq 2\|\G^\dagger - \G_\theta\|_{L^2_\mu(\A;\U)}\|\nabla_{\theta_i} \G_\theta\|_{L^2_\mu(\A;\U)}<\infty.
\]
We can then apply the dominated convergence theorem for Bochner integral and move the gradient inside the expectation in \cref{eq:SGD} to obtain
\begin{equation} \label{eq_approx_theta}
  \nabla_\theta (\E_{a\sim \mu}\|\G^\dagger(a) - \G_\theta(a)\|_{\U}^2) = \E_{a\sim \mu}[\nabla_\theta\|\G^\dagger(a) - \G_\theta(a)\|_{\U}^2].
\end{equation}

\paragraph{Function space hierarchy} For $1\leq i\leq m$, we define $\A_{i}$ to be a finite element space of functions defined on a mesh $\Omega_\A^{i}$ of $\Omega_\A$, with polynomial degree $k\geq d/2$, and mesh diameter $h_i=2^{-i}h_0>0$ such that $\dim(\A_1)\leq \cdots\leq \dim(\A_m)$ (note that one could use another refinement factor). For simplicity, we assume that $\A_i$ are subspaces of $\A$ so that the injection operator $\A_i\to\A$ is trivial. Let $X_\theta:\A\to \R$ be the operator defined as $X_\theta(a) = \|\G^\dagger(a) - \G_\theta(a)\|_{\U}^2$. We define the Galerkin projection operator $\P_{i}:\A\to \A_i$ and $X_\theta^i = X_\theta\circ \P_i$. Note that the neural operator $\G_\theta$ is discretization invariant following the framework introduced in~\cite{kovachki2023neural}, and that computing $X_\theta^1$ (and its gradient with respect to $\theta$) is inherently cheaper than computing $X_\theta^m$. Hence, the projection step to the latent dimension of the neural operator is computationally cheaper. Note that this applies for any neural operator architecture written in the encoder-processor-decoder framework~\cite{bouziani2024structure,KOVACHKI2024419}. FNOs~\cite{li2021fourier} could also be written in this framework by considering a spectral discretization of the function space, and the Fourier transform as the encoder.

\paragraph{Multi-Level Monte Carlo gradient estimation} The goal of the MLMC~\cite{giles2015multilevel} gradient estimation is to approximate the expectation of the gradient of the loss function with respect to the neural network parameters $\theta$ using a telescopic sum of differences across mesh levels. The main idea is to use a coarse mesh to capture most of the variance in the gradient, and then use finer meshes to correct for the remaining variance. This allows us to reduce the number of fine-mesh samples needed, thus speeding up the training process.
We express the fine-mesh gradient expectation, approximating $\E_{a\sim\mu}[\nabla_\theta X_\theta(a)]$, as a telescopic sum of differences across mesh levels and decompose the expectation of the gradient in \cref{eq_approx_theta} as follows:
\begin{equation} \label{eq_MLMC}
  \begin{aligned}
    \E_{a\sim\mu}[\nabla_\theta X_\theta^m(a)] & = \E_{a\sim\mu}[\nabla_\theta X_\theta^1(a)] + \sum_{i=2}^{m} \E_{a\sim\mu}[\nabla_\theta X_\theta^{i}(a)-\nabla_\theta X_\theta^{i-1}(a)]                                                                                                               \\
                                               & \approx \underbrace{\frac{1}{N_1}\sum_{j=1}^{N_1}[\nabla_\theta X_\theta^1(a_{1,j})]}_{Y_1} + \sum_{i=2}^{m} \underbrace{\left(\frac{1}{N_i}\sum_{j=1}^{N_{i}}[\nabla_\theta X_\theta^{i}(a_{i,j})-\nabla_\theta X_\theta^{i-1}(a_{i,j})]\right)}_{Y_i},
  \end{aligned}
\end{equation}
where $\{N_i\}_{1\leq i\leq m}$ denotes the sample sizes at each level. Note that we do not consider discretization of the output space $\mathcal{U}$ for simplicity, but a similar argument would also work, see, e.g., \cite{bouziani2024structure}.

\paragraph{Optimal allocation of samples} Following \cite[Sec.~1.3]{giles2015multilevel}, the sequence $Y_1+\ldots+Y_m$ in the second line of \cref{eq_MLMC} provides an unbiased estimator for $\E_{a\sim\mu}[\nabla_\theta X_\theta(a)]$, which approximates it with increasing accuracy, but also increasing cost, as we require more samples at the finest level of the mesh hierarchy. If we assume that each term inside the $Y_i$ can be evaluated at a computational cost of $C_i$ (where $C_1\leq \ldots\leq C_m$ as the mesh becomes finer), then the total cost of the estimator is
\[C = \sum_{i=1}^m N_i C_i.\]
We define the mean square error (MSE) of the estimator as
\[\textup{MSE} = \E_{a_{i,j}\sim \mu}\left[\left(\sum_{i=1}^m Y_i-\E_{a\sim\mu}[\nabla_\theta X_\theta(a)]\right)^2\right],\]
and aim to find an optimal allocation of samples $N_1,\ldots,N_m$ that minimizes the total computational cost, while achieving a mean squared error of $\epsilon^2$, for some $\epsilon>0$. The following proposition relies on a standard MLMC result~\cite{giles2008multilevel,giles2009multilevel,giles2015multilevel} and provides a mean squared error bound for the MLMC estimator in \cref{eq_MLMC}.

\begin{proposition}[Sample allocation for gradient estimation] \label{prop:MLMC}
  Let $\epsilon>0$ and suppose that the cost $C_i$ of each sample in \cref{eq_MLMC} scales linearly with respect to the number of degrees of freedom in the finite element space as $C_i \leq c 2^{di}$. Under \cref{ass_reg_G}, there exist a number of levels $m>0$ and an optimal allocation of samples $N_i\propto 2^{-(2k+d)i/2}$ per level such that the MLMC estimator to $\E_{a\sim\mu}[\nabla_\theta X_\theta(a)]$ given by \cref{eq_MLMC} has a mean squared error bounded by $\epsilon^2$.
\end{proposition}

\begin{proof}
  Let $1\leq i\leq m$ be the level of the mesh. For a given $a\sim \mu$, we are interested in bounding the difference between the gradients on the coarse and fine meshes in \cref{eq_MLMC}. We remark that $X_\theta^i(a) = X_\theta(\mathcal{P}_i a)$, and for $a\sim\mu$ denote $H_\theta(a) \coloneqq \G^\dagger(a) - \G_\theta(a)$. Since $\G^\dagger$ and $\G_\theta$ are Lipschitz continuous, the application $H_\theta:\A\to\U$ is Lipschitz continuous with Lipschitz constant $\mathrm{Lip}(H_\theta) = \mathrm{Lip}(\G^\dagger) + \mathrm{Lip}(\G_\theta)$. Let $1\leq s\leq p$ and $a\sim\mu$, then
  \begin{align*}
    |\nabla_{\theta_s} X_\theta(\mathcal{P}_i a)-\nabla_{\theta_s} X_\theta(a)| & = 2|\langle H_\theta(\mathcal{P}_i a),\nabla_{\theta_s} \G_\theta(\mathcal{P}_i a)\rangle_{\U}-\langle H_\theta(a),\nabla_{\theta_s}\G_\theta(a)\rangle_{\U}|                    \\
                                                                                & \leq 2|\langle H_\theta(a),\nabla_{\theta_s} \G_\theta(a)-\nabla_{\theta_s} \G_\theta(\mathcal{P}_i a)\rangle_{\U}|                                                              \\                                &\quad     + 2|\langle H_\theta(a)-H_\theta(\mathcal{P}_i a),\nabla_{\theta_s} \G_\theta(\mathcal{P}_i a)\rangle_{\U}|                                        \\
                                                                                & \leq 2 (L_\theta \|H_\theta(a)\|_\U                                         +\mathrm{Lip}(H_\theta)\|\nabla_{\theta_s}\G_\theta(\mathcal{P}_i a)\|_\U)\|a-\mathcal{P}_i a\|_{\A} \\
                                                                                & \leq C_\theta\|a-\mathcal{P}_i a\|_{\A},
  \end{align*}
  where we used the Cauchy-Schwarz inequality to obtain the second inequality, and introduced the constant $C_\theta>0$ to bound the first term in the second inequality uniformly over $a$ because $\mu$ is compactly supported on $H^k(\Omega_\A)$ and the operators are continuous. Here, we require that $\nabla_\theta\G_\theta$, $\G^\dagger$, and $\G_\theta$ are Lipschitz continuous, and use the fact that $a$ belongs to a compact domain and $H_\theta$, $\nabla_\theta \G_\theta$ are continuous to bound the last two terms uniformly over $a\sim \mu$. Finally, we employ a standard finite element error estimate~\cite[Eq.~4.4.28]{brenner2008mathematical} to bound the difference between the projection $\mathcal{P}_i a$ and the original function $a$ as
  \[\|\mathcal{P}_i a-a\|_{L^2(\Omega_\A)}\leq C 2^{-k i}\|a\|_{H^k(\Omega_\A)}.\]
  Using the compactness of the measure $\mu$, we conclude that there exists a constant $M>0$ independent of $i$ such that
  \begin{equation} \label{eq_bounded_exp}
    \|\nabla_{\theta} X_\theta^i(a)-\nabla_{\theta} X_\theta(a)\|_{2} \leq M 2^{-k i}.
  \end{equation}
  Moreover, by triangular inequality, we have
  \begin{equation} \label{eq_bounded_var}
    \|\nabla_{\theta} X_\theta^i(a)-\nabla_{\theta} X_\theta^{i-1}(a)\|_{2} \leq \|\nabla_{\theta} X_\theta^i(a)-\nabla_{\theta} X_\theta(a)\|_{2}+\|\nabla_{\theta} X_\theta^{i-1}(a)-\nabla_{\theta} X_\theta(a)\|_{2} \leq (1+2^k)M 2^{-k i}.
  \end{equation}
  Combining \cref{eq_bounded_exp,eq_bounded_var}, we obtain the following bound for the expectation of the approximation of the gradient, and the variance of the multi-level estimator:
  \[
    \mathbb{E}_{a\sim \mu}[\|\nabla_{\theta} X_\theta^i(a)-\nabla_{\theta} X_\theta(a)\|_{2}] \leq M 2^{-k i},             \quad
    \textrm{Var}_{a\sim \mu}[\|\nabla_{\theta} X_\theta^i(a)-\nabla_{\theta} X_\theta^{i-1}(a)\|_2]  \leq \frac{(1+2^k)^2 M^2}{4}2^{-2ki},
  \]
  where we used Popoviciu's inequality to bound the variance of the bounded random variable $\|\nabla_{\theta} X_\theta^i(a)-\nabla_{\theta} X_\theta^{i-1}(a)\|_2$. Moreover, the assumption on the growth of the computational cost with respect to the number of levels implies that $C_i\leq c 2^{di}$. Then, following~\cite[Thm.~2.1]{giles2015multilevel}, for any $\epsilon>0$, there exists a number of level $m$, and optimal allocation of samples $N_i\propto 2^{-d i/2}$ per level such that the MLMC estimator given by \cref{eq_MLMC} has a mean square error bounded by $\epsilon^2$.
\end{proof}

Following \cref{prop:MLMC}, we find that one can reduce the number of fine-mesh samples to accurately estimate the gradient of the loss function by adapting the number of samples at each level depending on the mesh resolution. Note that the number of samples per resolution is related to the sample cost assumption, which depends on the type of neural operator architecture. The analysis can also be generalized to non-geometric hierarchy of meshes~\cite{haji2016optimization}.

\begin{remark} (Multi-Level Quasi-Monte Carlo) \label{rem:MLMC_QMC}
  The computational cost of MLMC can be reduced by employing a Quasi-Monte Carlo (QMC) approach, where a nested set of samples across mesh resolutions is chosen instead of independent realizations at each level~\cite{kuo2012quasi,kuo2015multi}. This increases the variance of the estimator but results in a lower cost complexity~\cite{giles2015multilevel}.
\end{remark}

\subsection{Mini-batch strategies} \label{sec_mini_batch}
The mathematical framework presented in \cref{sec:math_fram} is based on the Gradient Descent (GD) algorithm for training neural networks. However, mini-batching is commonly used in practice due to its advantages in computational efficiency, generalization performance, and scalability. In this context, the MLMC gradient estimation (see \cref{eq_MLMC}), which adjusts the amount of samples across resolution levels, naturally extends to mini-batch training and reads as follows:
\begin{equation}\label{eq:MLMC_batch}
  \mathbb{E}_{a\sim\mu}[\nabla_\theta X_\theta^m(a)] \approx \underbrace{\frac{1}{B_1}\sum_{j=1}^{B_1}[\nabla_\theta X_\theta^1(a_{1,j})]}_{\textnormal{coarsening loss}} + \sum_{i=2}^{m} \underbrace{\left(\frac{1}{B_i}\sum_{j=1}^{B_{i}}[\nabla_\theta X_\theta^{i}(a_{i,j})-\nabla_\theta X_\theta^{i-1}(a_{i,j})]\right)}_{\textnormal{pairing losses}},
\end{equation}
where $B_i$ stands for the batch size at the {correction level} $i$ in the sum hierarchy, obtained by drawing samples from a subset of $N_i$ and $N_{i-1}$ observations at resolution $R_i$ and $R_{i-1}$, respectively, for $2\leq i \leq m$. The term $\nabla_\theta X_{\theta}^{i}(a_{i,j}) - \nabla_\theta X_{\theta}^{i-1}(a_{i,j})$ in Eq.~(6) is therefore the \textit{gradient correction at level $i$}, i.e., the difference between gradients evaluated on two consecutive discretization levels $i$ and $i - 1$, for $2\leq i \leq m$. For clarity, throughout the paper we use level $i$ to indicate both the mesh resolution index and its associated gradient correction in the MLMC estimator. Moreover, the term $-\nabla_\theta X_\theta^{i-1}(a_{i,j})$ in \cref{eq:MLMC_batch} is computed by coarsening the sample $a_{i,j}$ to resolution $R_{i-1}$ and evaluating the neural operator. Following \cref{sec:math_fram}, we decrease $B_i$ and $N_i$ with respect to the level $i$ as the data resolution $R_i$ increases.

\paragraph{Sample and batch sizes} We denote by $\mathcal{N}_i=\{(a_{i,j},u_{i,j})\}_j$ the set of data observations at level $i$, and by $N$ the total number of samples. We determine the evolution of the number of samples $N_i = |\mathcal{N}_i|$ according to one of the following approaches:
\begin{itemize}[leftmargin=*,topsep=0pt]
  \item \textit{Geometric progression.} There is a constant growing factor $\delta\geq 1$ between a level $i$ and a subsequent level $i+1$ as $N_{i} = \delta N_{i + 1}$, for $1\leq i \leq m - 1$, such that $\sum_{i=1}^m N_i \approx N$.
  \item \textit{Prescribed progression.} We manually select the amount of samples at level $i$ as $N_{i}$, such that $N_m\leq \ldots\leq N_1$ and $\sum_{i=1}^m N_i = N$.
  \item \textit{Optimal progression.} The evolution of samples follows the ``optimal'' geometric progression $N_{i} = 2^{-(2k+d)i/2} N_{i - 1}$, for $i = 2, \ldots, m$, determined by \cref{prop:MLMC}.
\end{itemize}
Note that while \cref{prop:MLMC} provides an optimal choice for the allocation of samples, the geometric factor intrinsically depends on the sampling cost assumption, which is architecture-dependent. We will compare the different sampling strategies described above in different numerical experiments in \cref{sec_experiments}. The batch sizes $B_m\leq \ldots \leq B_1$ are defined through a geometric progression of the form $B_{i} = \delta B_{i + 1} = \delta^{m - i} B_m $, where the batch size $B_m$ at the last level and the multiplier $\delta \ge 1$ are prescribed manually.

\paragraph{Sampling strategy} Data sampling is handled through a \texttt{batcher} providing the data structure $\smash{\mathcal{B}=[\mathcal{B}^1, \ldots, \mathcal{B}^{N_i/ B_i}]}$, implemented as a list of $N_i/ B_i$ \texttt{batch} objects for $1\leq i \leq m$. Each \texttt{batch} $\smash{\mathcal{B}^k = [\mathcal{B}_1^k, \ldots \mathcal{B}_m^k]}$ is itself a list of length $m$, where $\smash{\mathcal{B}_i^k}$ represents the set of data indices at correction level $i$ and resolution $R_i$. The first element of the \texttt{batch}, i.e., $\smash{\mathcal{B}_1^k}$, is a set of $B_1$ indices of coarse data sampled uniformly from $\mathcal{N}_1$. We explore two strategies to select the finer data indices within the $k$-th level of the \texttt{batch}:
\begin{itemize}[leftmargin=*,topsep=0pt]
  \item \textit{Random sampling.} We uniformly sample $\smash{\mathcal{B}_{2}^k,\ldots \mathcal{B}_{m}^k}$ from the datasets $\smash{\mathcal{N}_2,\ldots,\mathcal{N}_m}$. This is a standard multi-level Monte Carlo sampling and ensures independence of the data samples across levels in the telescopic sum given by \cref{eq:MLMC_batch}. We repeat the sampling procedure for every epoch to ensure full coverage of the data.
  \item \textit{Nested sub-sampling.} To increase computational efficiency, data indices are sampled uniformly from the previous level in the resolution hierarchy as $\mathcal{B}_{i}^k \subset \mathcal{B}_{i-1}^k$, for $2\leq i \leq m$. In contrast to random sampling, nested sub-sampling does not produce independent samples at each level. This sampling strategy can be viewed as a Multi-Level Quasi-Monte Carlo method (cf.~\cref{rem:MLMC_QMC}). We repeat the sampling procedure at every epoch to ensure full coverage of the data.
\end{itemize}

\begin{figure}[htbp]
  \centering
  \includegraphics[width=0.6\textwidth]{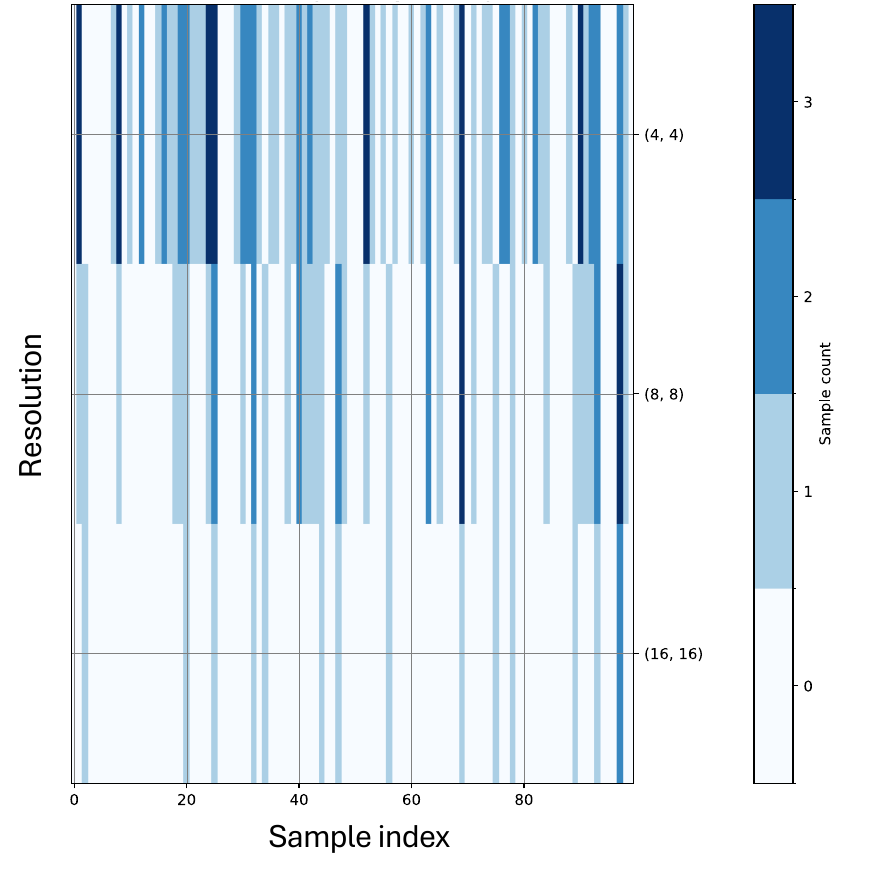}
  \caption{Illustration of random sampling strategy in the construction of $\mathcal{B}$ for $m=3$ levels.}
  \label{fig:random_sampling}
\end{figure}

\begin{figure}[htbp]
  \centering
  \includegraphics[width=0.9\textwidth]{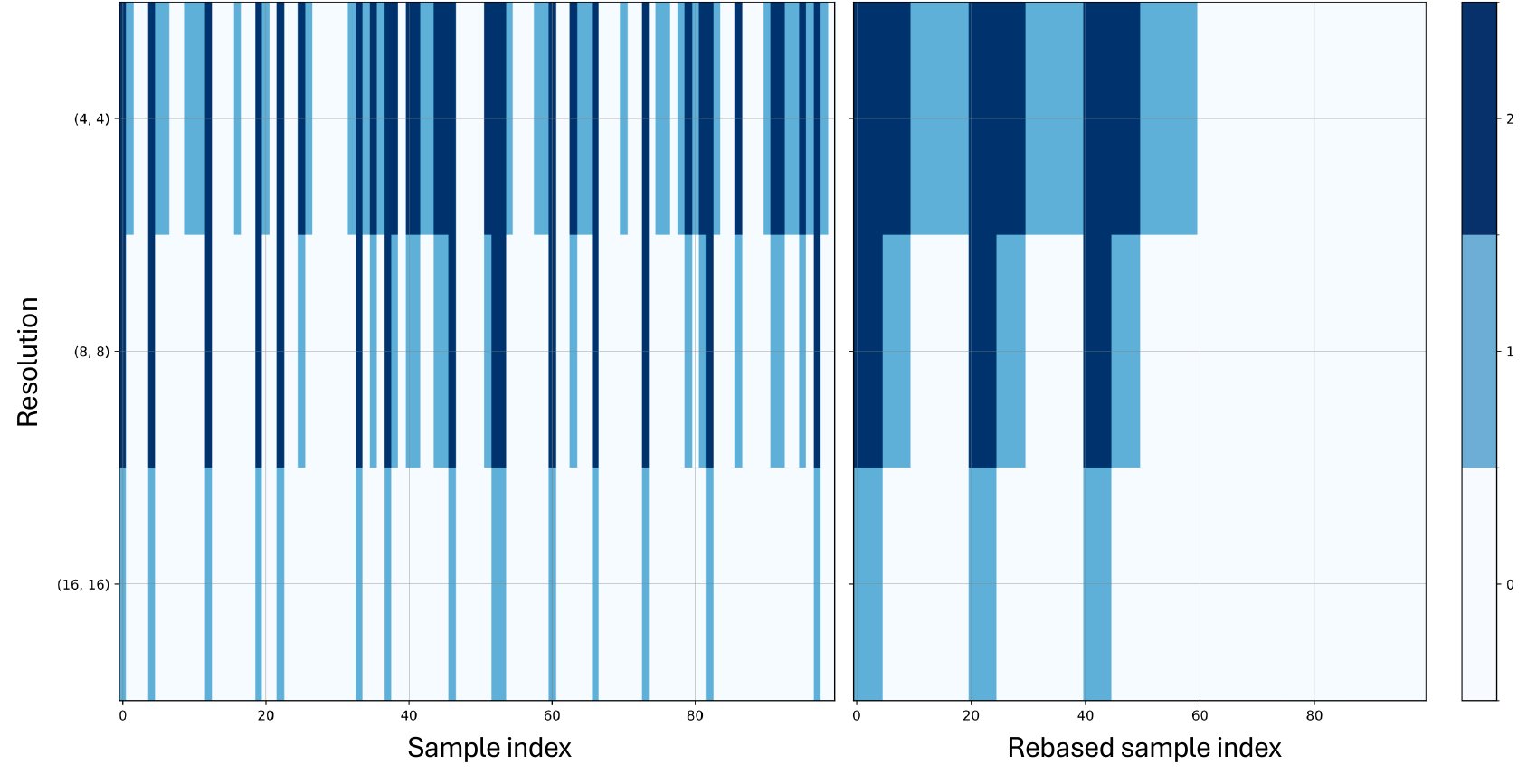}
  \caption{Illustration of nested sub-sampling strategy in the construction of $\mathcal{B}$ for $m=3$ levels. }
  \label{fig:sawtooth_plot}
\end{figure}

In \cref{fig:random_sampling,fig:sawtooth_plot} we report an illustration of the random sampling and nested sub-sampling strategies.  In particular, we consider $N = 105$ samples, $m=3$ levels ($R_1 = 4 \times 4, R_2 = 8 \times 8, R_3 = 16 \times 16$), a multiplier $\delta = 2$ and the geometrical progression. Data samples are  selected at each level $i$, i.e. resolution, by obtaining the sets $\mathcal{N}_1, \mathcal{N}_2, \mathcal{N}_3$ of dimension $N_1 = 60, N_2 = 30, N_3 = 15$ and batches $\mathcal{B}_{1}^k, \mathcal{B}_{2}^k, \mathcal{B}_{3}^k$, $k = 1, \ldots, 3$, of dimension $B_1 = 20, B_2 = 10, B_3 = 5$.

\cref{fig:random_sampling} illustrates the random sampling strategy, where the finer data indices are uniformly sampled from the original dataset. Note that for each resolution, it is possible to select samples that are not considered in the batches referring to a different resolution.  The sample count refers to the number of times a sample is present in $\mathcal{B}$, not only because it can be sampled in another batch with the same resolution, but also because it can be chosen as a sample to be used in a correction term. To give some intuition, we start by uniformly sampling the indices of data related to $R_1 = 4 \times 4$ and belonging to $\mathcal{B}^k_1$, with $k = 1, \ldots, 3$, each of dimension $B_1 = 20$. Then, we proceed be selecting the indices for the gradient correction at level $i=2$, that is batches $\mathcal{B}^k_2$, with $k = 1, \ldots, 3$, with batch size $B_2 = 10$. The same indices must also be selected at resolution $R_1$ in order to be able to compute the gradient correction term. The mechanism in then iterated to compute the gradient correction term at level $i=3$.

In \cref{fig:sawtooth_plot} finer data are sub-sampled from the previous resolution indices with the nested sub-sampling technique. In \cref{fig:sawtooth_plot} (left), we show the selections of data indices at different resolutions; it is not possible to have a sample at level $i = 3$ that has not also been selected at previous levels $i = 1, 2$. After obtaining the selected data indices, we perform a rebasing operation to better show the structure of batches $\mathcal{B}_{1}^k, \mathcal{B}_{2}^k, \mathcal{B}_{3}^k$, $k = 1, \ldots, 3$, resulting from nested sub-sampling in \cref{fig:sawtooth_plot} (right).

\paragraph{Multi-resolution data batcher}
To preserve the computational efficiency of MLMC, we implement a cache-aware data management and loss computation system that optimizing memory transfers between host and device. Following \cref{alg:MLMC-SGD}, the data \texttt{batcher} pre-loads the \texttt{batch} structures $\mathcal{B}^k$ onto GPU memory at the beginning of each epoch and implements a resolution-aware memory manager that maintains a memory layout for tensors across resolution levels, facilitating efficient correction term computations. This approach achieves up to $\mathcal{O}(m)$ reduction in transfer overhead compared to naive implementations, while maintaining a bounded GPU memory usage of $\mathcal{O}(\sum_{i=1}^m B_i R_i^d)$.

\subsection{MLMC optimization algorithm}

Our MLMC optimization algorithm is described in \cref{alg:MLMC-SGD} and implements the mini-batch strategies introduced in \cref{sec_mini_batch}, along with the telescopic sum evaluation of the gradient given by \cref{eq:MLMC_batch}. At a given gradient correction level $2\leq i\leq m$, the loss is evaluated on samples $(a_{i,j},u_{i,j})$ from the batch $\mathcal{B}_i^k$ of the observation set $\mathcal{N}_i$ at resolution $R_i$. We then subtract the loss value at the same samples but coarsened to resolution $R_{i-1}$. The set of corresponding coarsen samples is denoted by $\tilde{\mathcal{N}}_{i-1}$. Note that it is essential to maintain consistency between the samples at different resolutions within a given correction level $i$ when constructing the data \texttt{batcher}, in order to preserve the validity of the MLMC estimator and compute the correct pairing losses in \cref{eq:MLMC_batch}. Once the gradient has been computed, any gradient-based optimization algorithm can be used to update the parameters of the neural operator, such as Adam~\cite{kingma2014adam}.

\begin{center}
  \begin{algorithm}[t]
    \caption{MLMC optimization algorithm}
    \label{alg:MLMC-SGD}
    \begin{algorithmic}[1]
      \STATE {\bf Input}: Number of levels $m$, observations $\mathcal{N}_i = \{(a_{i,j},u_{i,j})\}_{j=1}^{N_i}$ at resolution $R_i$, batches $\mathcal{B}_i^k$, batch sizes $B_i$, sampling strategy, number of epochs $n_e$, optimizer
      \STATE {\bf Output}: $\theta^*$
      \STATE Initialize $\theta_0$
      \STATE Initialize \texttt{batcher}
      \FOR{$e$ in $n_e$}
      \STATE $\mathcal{B}$ = \texttt{batcher}.sample($m$, $\mathcal{N}_i$, $B_i$, sampling strategy)
      \FOR{$\mathcal{B}^k$ in $\mathcal{B}$}
      \STATE $\mathtt{total\_loss} = \mathtt{loss}(\mathcal{N}_1, \mathcal{B}_1^k)$
      \FOR{$i$ in $\{2,\ldots, m\}$}
      \STATE $\mathtt{total\_loss} \pluseq \mathtt{loss}(\mathcal{N}_i, \mathcal{B}_i^k) - \mathtt{loss}(\tilde{\mathcal{N}}_{i-1}, \mathcal{B}_{i-1}^k)$
      \ENDFOR
      \STATE $\mathtt{gradient = total\_loss.backward()}$
      \ENDFOR
      \STATE $\mathtt{optimizer.step()}$
      \ENDFOR
    \end{algorithmic}
  \end{algorithm}
\end{center}

\section{Numerical experiments} \label{sec_experiments}
We evaluate the efficiency of the MLMC training algorithm for neural operators on three state-of-the-art neural operators architectures: \textit{(i)} Fourier neural operator (FNO) \cite{li2021fourier}, \textit{(ii)} message passing neural PDE solver (MP-PDE) \cite{brandstettermessage}, and \textit{(iii)} geometry-informed neural operator transformer (GINOT) \cite{liu2025geometryinformedneuraloperatortransformer}. Additional experiments with \textit{(iv)} mesh-free convolutional neural network \cite{zakariaei2025multiscale} are also reported. The test-cases considered in this work consist of standard (time-dependent) nonlinear PDE problems, and include a fluid flow past a cylinder, and a complex 3D jet engine brackets dataset~\cite{10.1115/1.4067089}. Note that the aim of this work is not to introduce new architectures, but rather improve the computational cost/accuracy trade-off in operator learning through MLMC training. All experiments are repeated on three random seeds and the results are averaged. The simulations are performed on a single Nvidia A100 GPU with 80GB of RAM. Code to reproduce all experiments in this study is openly available at \url{https://github.com/JRowbottomGit/mlmc_optim}.

\subsection{Fourier neural operator} \label{sec_fno}

FNOs are a family of neural operators that exploit spectral (Fourier) discretizations of the input and output functions~\cite{li2021fourier}. These neural operators accept data at different resolutions using projection onto a latent dimension of fewer Fourier modes, which are then processed by Fourier convolutions. The MLMC training procedure introduced in \cref{sec_method} decreases the computational cost of the initial and final spectral convolutional layers and allows us to use fewer high-resolution data. This may be critical in large-scale applications where the cost of data acquisition might be high.

\paragraph{Darcy flow problem}

We consider the steady state Darcy flow equation on the domain $\Omega = [0,1]^2$ as~\cite{li2021fourier}:
\begin{equation}   \label{eq:darcy}
  \begin{aligned}
    - \nabla \cdot (a(\mathbf{x}) \nabla u(\mathbf{x})) & = 1, \quad \mathbf{x}\in\Omega,                                                  \\
    u(\mathbf{x})                                       & = 0,                                       \quad \mathbf{x} \in \partial \Omega, \\
  \end{aligned}
\end{equation}
where the diffusion coefficient $a: \Omega \rightarrow \mathbb{R}$ is chosen such that $a(\mathbf{x}) = \psi\circ \mu(\mathbf{x})$, where $\mu \sim \mathcal{N}(0, (-\Delta + 9I)^{-2})$ with zero Neumann boundary conditions on the Laplacian, and $\psi:\mathbb{R}\to\mathbb{R}$ is defined as $\psi(x)=12$ if $x\geq 0$ and $3$ otherwise. We are interested in learning the nonlinear operator $a\mapsto u$ mapping the diffusion operator $a$ to the corresponding solution $u$ to \cref{eq:darcy}. The equation is solved numerically for $N=1024$ random diffusion coefficients on an equispaced $241\times 241$ grid using a second-order finite difference scheme, and downsampled to the lower resolutions such that $\{ R_i \times R_i \}_{i=1}^5= \{15 \times 15, 30 \times 30, 60 \times 60, 120 \times 120, 241 \times 241 \}$.

\begin{figure}[ht]
  \centering
  \includegraphics[width=\textwidth]{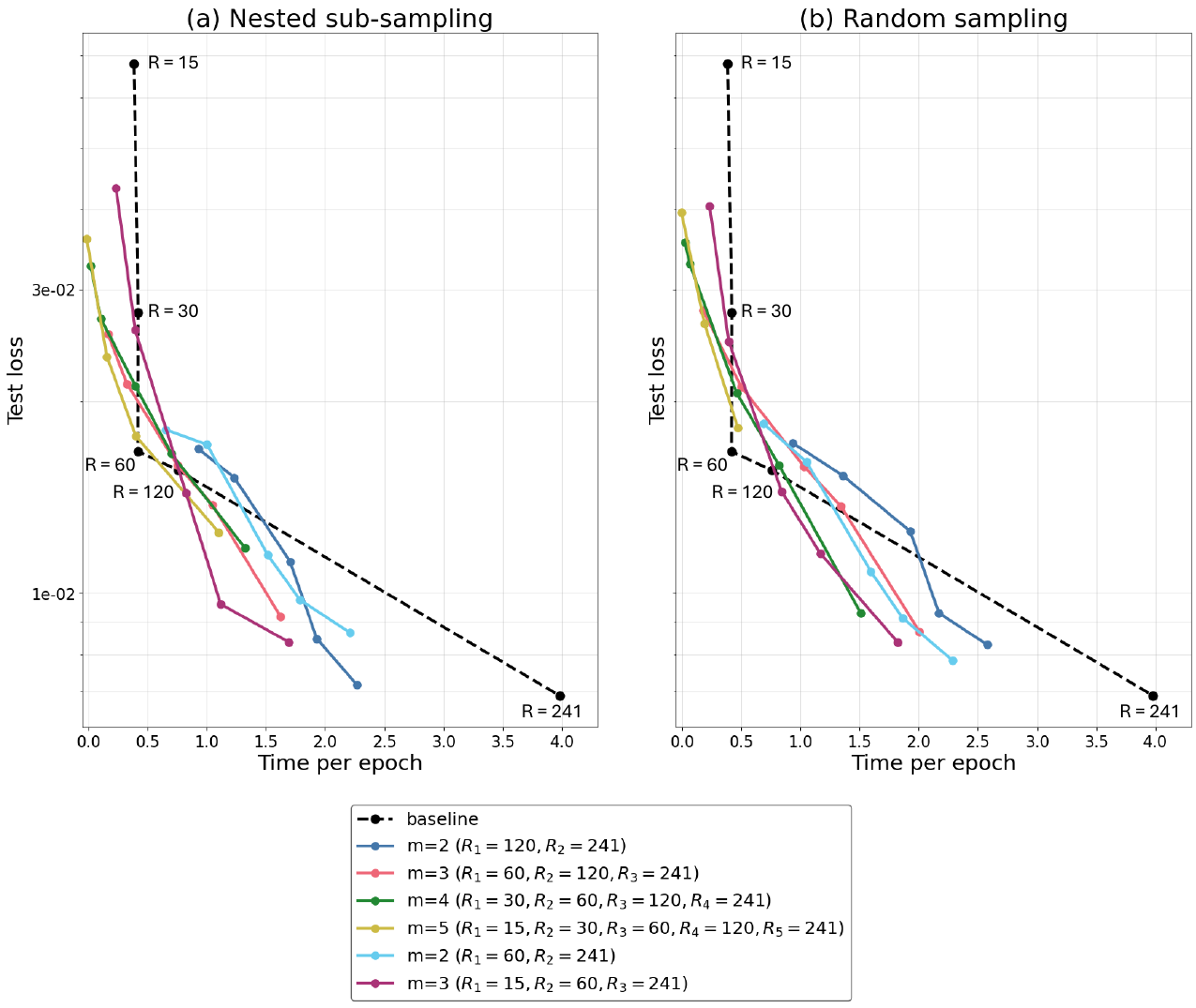}
  \caption{\textit{Darcy test problem.} Final test loss against the average computational training time per epoch. The baseline plotted as a dotted black line reports the standard FNO performance when trained on increasing data resolutions, while the colored lines highlight a ``Pareto curve'' of MLMC accuracy/training time trade-off when varying the number of levels $m$ and sampling factor $\delta$.}
  \label{fig:darcy_FNO_pareto_bound}
\end{figure}

To ensure fairness of the experiments, we keep the same hyper-parameters employed in the code released by \cite{li2021fourier}, and train the original FNO at different resolutions, which we refer to as the baseline below. We report in \cref{fig:darcy_FNO_pareto_bound} the final test accuracy against average (over 500 epochs) computational training time per epoch of the standard FNO baseline as the data resolution increases from $15\times 15$ to $\{30\times 30, 60\times 60,120\times 120,241\times 241\}$. We compare the results with the ``Pareto curve'' for different number of levels $m\in\{2,3,4,5\}$, resulting from MLMC training  with decreasing sampling multiplier factor $\delta\in\{8,4,2,1.5,1\}$. As $\delta$ decreases, a higher number of fine-resolution data are used, which increases the accuracy but also the computational time. The proposed MLMC training algorithm enables us to reach almost the same testing accuracy achieved by traditional training at $R = 241$, while requiring only $60\%$ of the training time. For completeness, we also report the tests referred to as ``skip'' in which the resolution scaling factor is $4$ (instead of $2$), and the considered resolutions are $15\times 15$, $60\times 60$, and $241\times 241$, as well as the nested sub-sampling strategy in \cref{fig:darcy_FNO_pareto_bound} (right). In this way, we further increase the computational performance by reducing the total amount of fine samples required.

\begin{figure}[t]
  \centering
  \includegraphics[width=\textwidth]{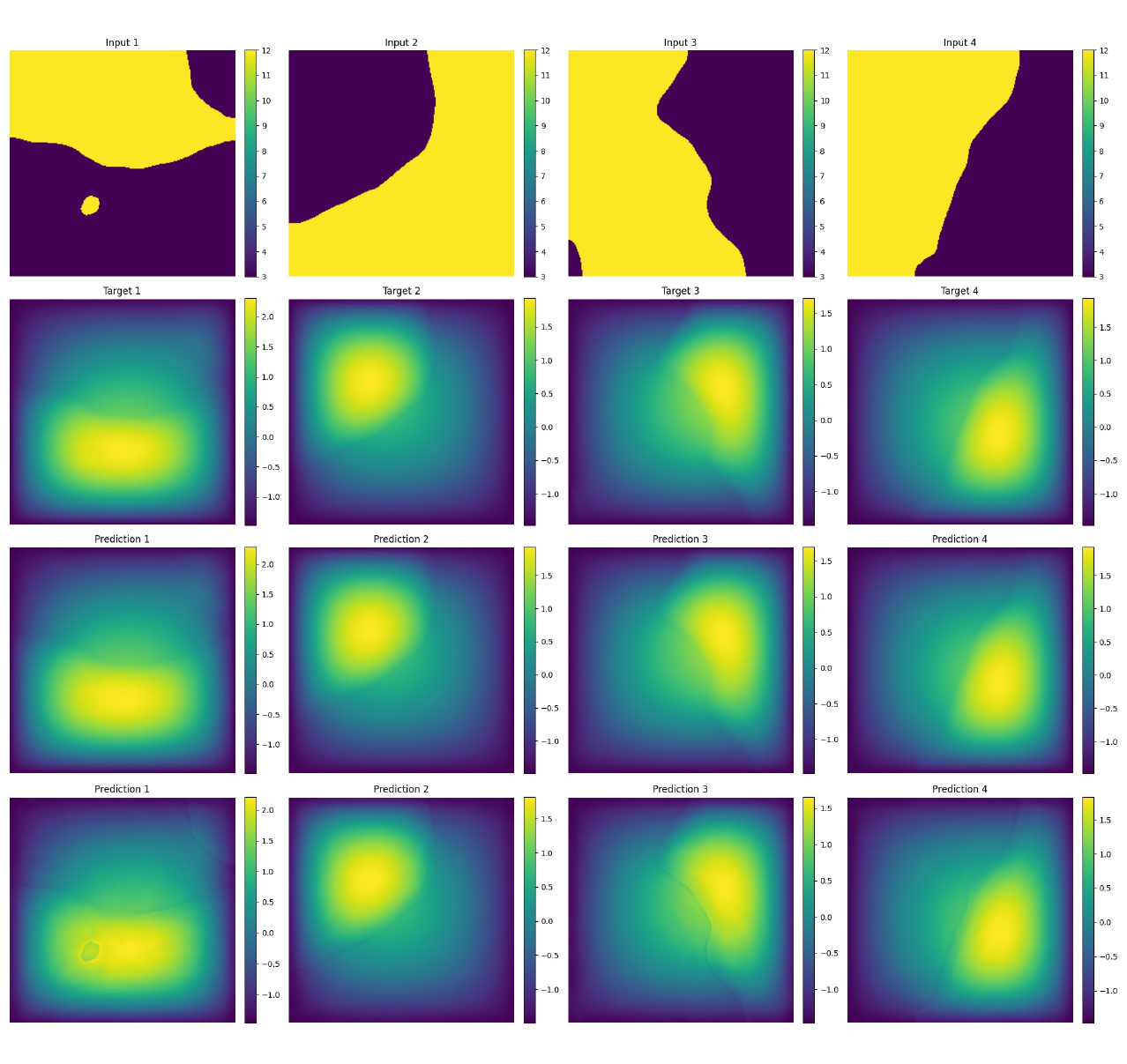}
  \caption{\textit{Darcy test problems.} Comparison between the ground truth data (second row), and the MLMC solutions with $m = 4$ and $\delta = 1$ (third row), and $m = 4$ and $\delta = 8$ (fourth row).}
  \label{fig:darcy_FNO_comparison_field}
\end{figure}

We compare the ground truth data with the FNO approximation obtained through the MLMC training algorithm in \cref{fig:darcy_FNO_comparison_field} by considering the values of $m, \delta$ giving the best performance in terms of accuracy and computational time, respectively. More precisely, the third row refers to the choice $m = 4$ and $\delta = 1$, using the random pairing strategy, while the fourth one to $m = 4$ and $\delta = 8$, always using random as pairing strategy. Regarding the latter, the error with respect to the ground truth data is about 3.56$\%$ and the epoch training time is 0.054$s$, leading to a speed-up almost equal to 8 with respect to the baseline with $R = 30$.

\begin{figure}[htbp]
  \centering
  \includegraphics[scale = 0.3]{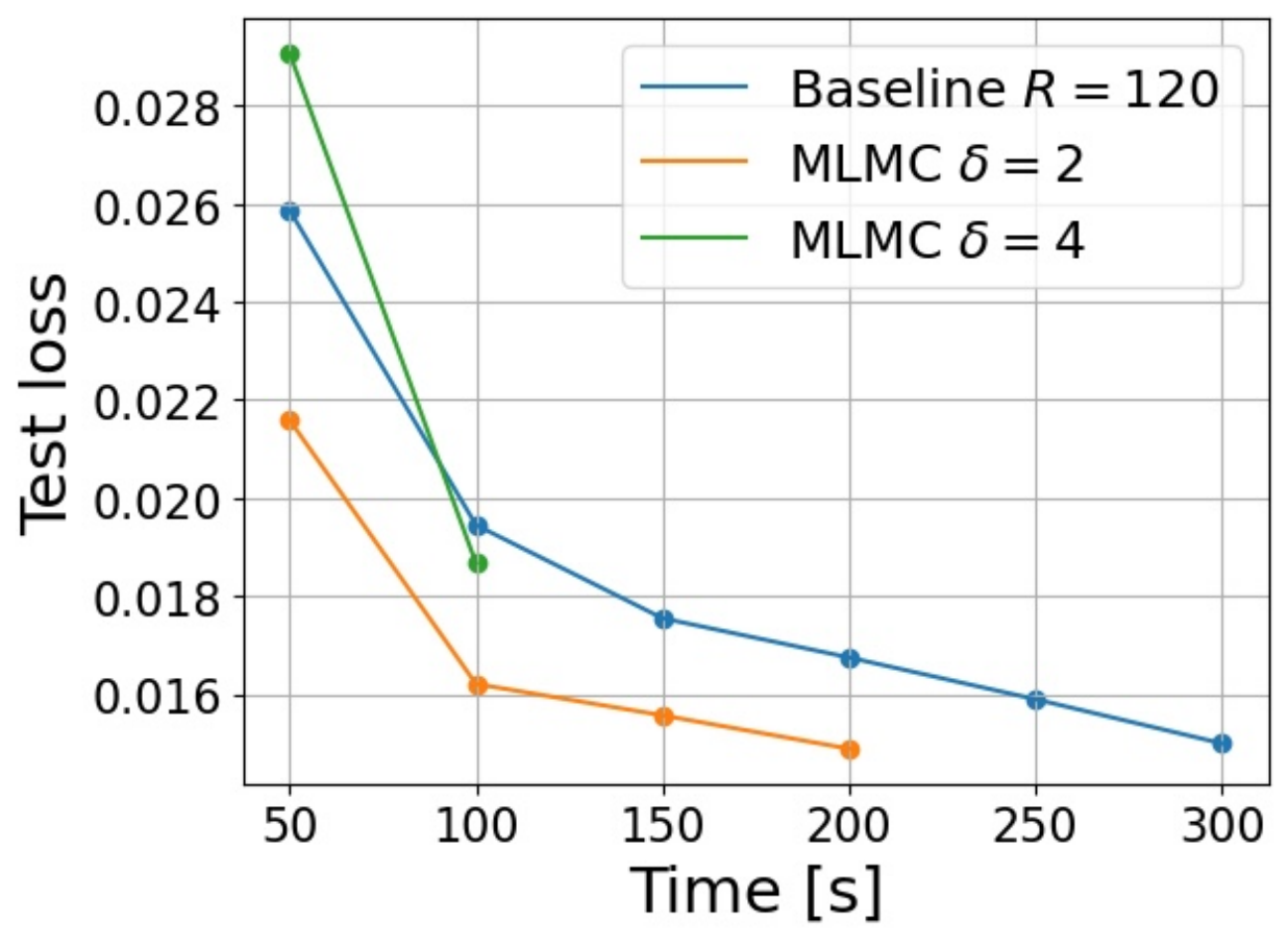}
  \caption{\textit{Darcy problem.} Final testing loss against cumulative training computational time.}
  \label{fig:darcy_res_gen}
\end{figure}

Moreover, we test the generalization capabilities of the FNO, trained by means of the MLMC algorithm, on the finest resolution $R = 241$, by varying the training budget. This analysis is motivated by the interest to evaluate the ability of the MLMC training algorithm to improve resolution generalization. We considered $m = 2$ levels, resolutions $\{ R_i \times R_i \}_{i = 1}^2 = \{ 30 \times 30$, $120 \times 120 \}$, resampling multiplier factor $\delta = 2,4$ and the nested sub-sampling technique to test the generalization performance on the finest resolution $R= 241$. The results are reported in \cref{fig:darcy_res_gen} where we report the final testing loss against the cumulative training computational time for the baseline with $R = 120$ and the MLMC training algorithm approximation. The MLMC training algorithm enables to improve resolution generalization by achieving a prescribed accuracy for a lower computational budget.

Finally, we compare the training computational cost/accuracy trade-off enabled by the MLMC training algorithm employing the gradient estimation formula in \cref{eq_MLMC} with respect to the one obtained by means of a mixed-resolution batches training without telescoping corrections. In particular, we perform the test by using the training specifics $m=4$ and $\delta=2$. The average training computational time obtained by the mixed-resolution batches approach is $1.13s$ with a final testing loss equal to $0.022$. In contrast, with the MLMC training algorithm we obtain the same accuracy but requireing just the $37\%$ of training computational time.

\paragraph{Time-dependent Navier-Stokes equations}
We then consider the approximation of the time-forward solution operator associated to the Navier-Stokes equations using a 3D Fourier neural operator (FNO-3d) that directly convolves in space-time. The two-dimensional time-dependent Navier-Stokes equations are expressed in vorticity form on the unit torus as follows~\cite{li2021fourier}:
\begin{equation}
  \begin{aligned}
    \frac{\partial {w}(\mathbf{x}, t)}{\partial t} + \mathbf{u}(\mathbf{x}, t) \cdot \nabla {w}(\mathbf{x}) - \nu \Delta w(\mathbf{x}, t) & =  f(\mathbf{x}, t), &  & (\mathbf{x},t) \in (0,1)^2 \times [0,T], \\
    \nabla \cdot \mathbf{u}(\mathbf{x}, t)                                                                                                & = 0,                 &  & (\mathbf{x},t) \in (0,1)^2 \times [0,T], \\
    w(\mathbf{x}, 0)                                                                                                                      & = w_0(\mathbf{x}),   &  & \mathbf{x} \in (0,1)^2,                  \\
  \end{aligned}
  \label{eq:NS}
\end{equation}
where $\mathbf{u}: \Omega \times [0, T) \rightarrow \mathbb{R}^2$ is the velocity field and $w = \nabla \times \mathbf{u}$ is the vorticity field. The initial condition for the vorticity field is generated as $w_{0} \sim\mathcal{N}\bigl(0,\,7^{3/2}(-\Delta + 49I)^{-2.5}\bigr)$, the forcing term is given by $f = 0.1(\sin(2 \pi(x+y)) + \cos(2 \pi(x+y)))$, and the viscosity coefficient is equal to $\nu = 10^{-3}$.  \cref{eq:NS} are equipped with periodic boundary conditions. We are interested in learning the operator which maps the vorticity field up to time $t=10$ to the vorticity up to time $T=50$. The equations are numerically solved by a pseudo-spectral method and a Crank--Nicolson time scheme for the viscous term with an explicit treatment of the nonlinear term on a uniform grid of resolution $256\times 256$, and a time discretization step $\Delta t = 10^{-4}$. The dataset is generated by downsampling the solutions to $\{ R_i \times R_i \}_{i = 1}^4 = \{ 8 \times 8$, $16 \times 16$, $32 \times 32, 64 \times 64\}$ space discretization points and storing them at every time units $t\in \{1,\ldots,50\}$. The number of samples of the initial conditions $\omega_0$ and resulting trajectories is set to $N = 3000$.

\begin{figure}[htbp]
  \centering
  \includegraphics[width=\textwidth]{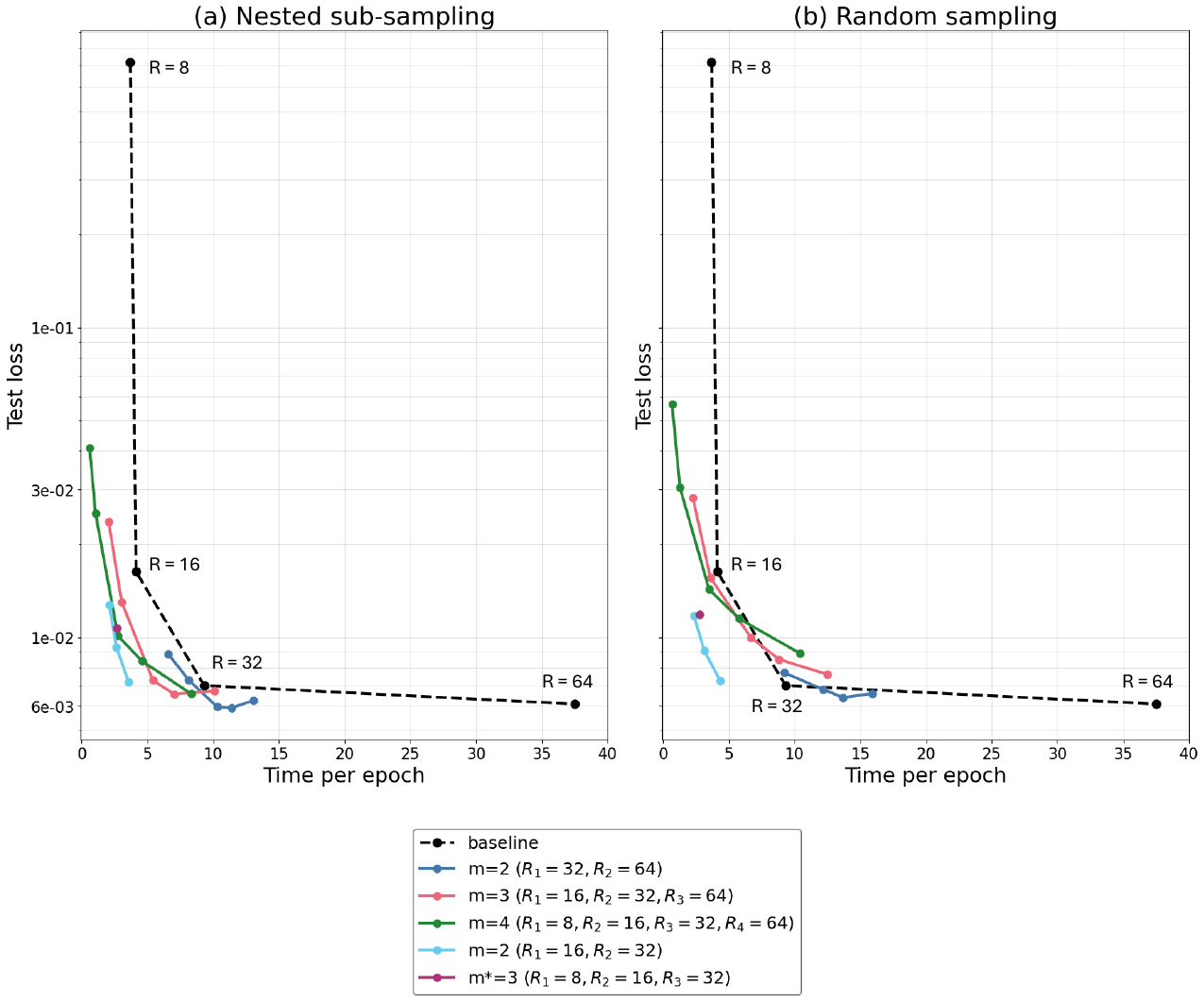}
  \caption{\textit{Navier-Stokes problem.} Same as \cref{fig:darcy_FNO_pareto_bound} with an FNO-3d trained on the time-dependent Navier-Stokes equations at increasing mesh resolutions $8\times 8$, $16\times 16$, $32\times 32$, $64\times 64$ (baseline) or with MLMC. The trained models are evaluated on a distinct test set at the finest resolution $64\times 64$. The magenta curve corresponds to MLMC at finest level $32\times 32$ and coarsest level $16\times 16$ with decreasing $\delta$, and yields a better cost/accuracy trade-off compared to the baseline.}
  \label{fig:NS_fno3d_pareto_bound}
\end{figure}

\begin{figure}[htbp]
  \centering
  \begin{overpic}[height=0.35\textwidth, trim=0 0 155 140,clip]{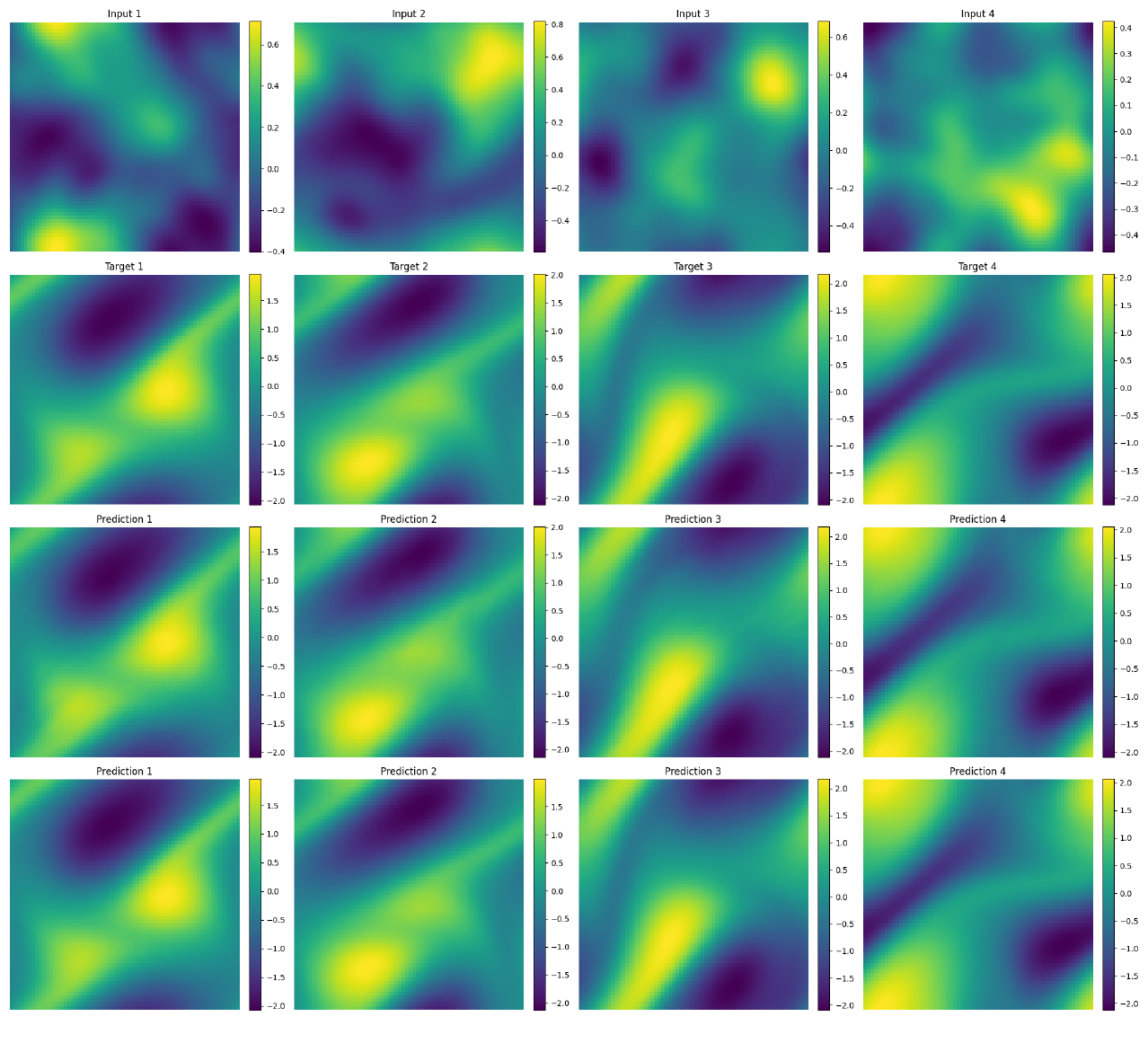}
    \put(-5,66.6){\rotatebox{90}{\tiny Ground truth}}
    \put(-5,37){\rotatebox{90}{\tiny MLMC (4,4)}}
    \put(-5,8){\rotatebox{90}{\tiny MLMC (2,2)}}
  \end{overpic}
  \hspace{0.05\textwidth}
  \includegraphics[height=0.35\textwidth]{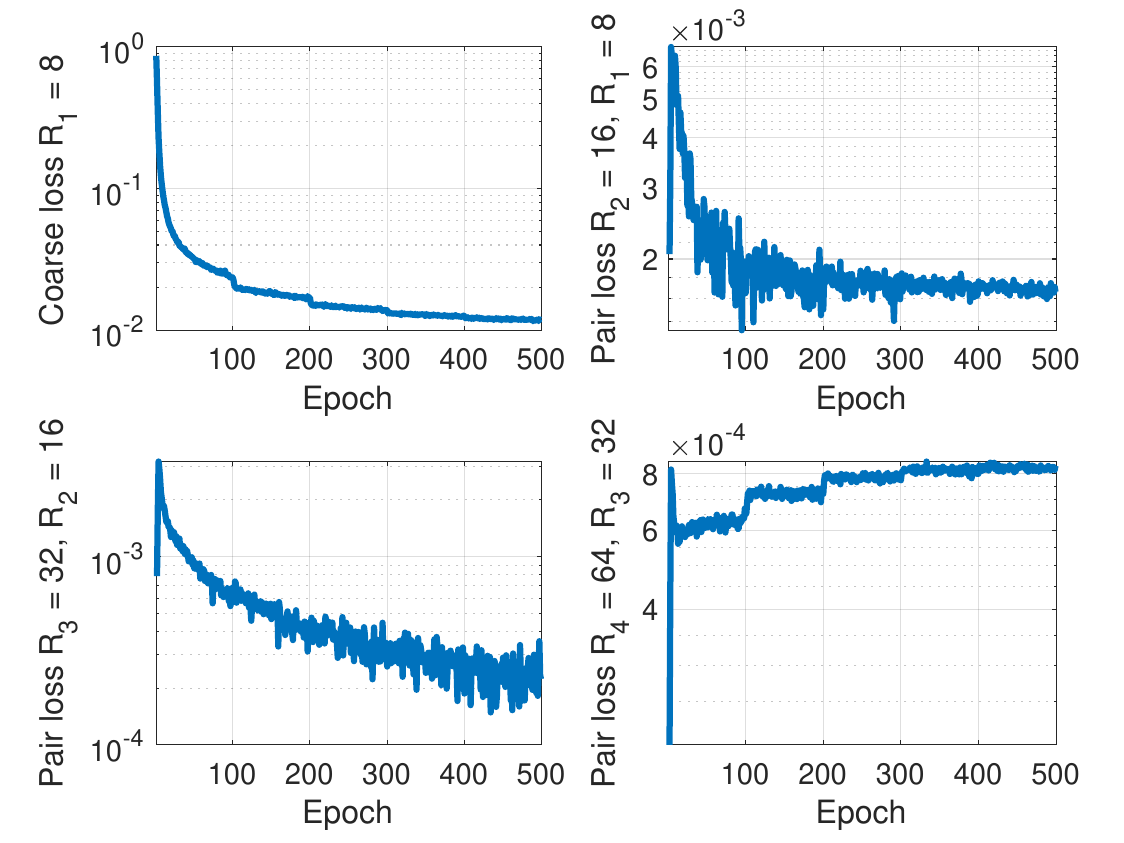}
  \caption{\textit{Navier-Stokes problem.} Left: Comparison between the ground truth data (1st row), and the MLMC predicted solutions at $t = 40$ with $(m,\delta) = (4,4)$ (2nd row) and $(2,2)$ (3rd row) for three different initial conditions. Right: Evolution of the coarse loss at the lowest resolution during training (top-left) and the pair losses in the telescopic sum expansion of the fine loss with $\delta=2$.}
  \label{fig:NS_fno3d_plot_best_worst}
\end{figure}

Again, we employ the same hyper-parameters employed in the code released by \cite{li2021fourier}. In \cref{fig:NS_fno3d_pareto_bound}, we display the Pareto frontier, in terms of average epoch training time and final testing loss, obtained by considering the standard FNO-3d training at different mesh resolutions (the baseline), and MLMC training with the two sampling strategies considered. Focusing on the baselines, we note that the testing loss values obtained at training resolutions $32\times 32$ and $64\times 64$ are very similar. Following this observation, we apply MLMC training with $m=2$ levels at finest resolution of $32\times 32$ and coarsest resolution of $16\times 16$, while decreasing the sampling factor $\delta$ from $8,4,2$. The result is illustrated as a cyan curve in \cref{fig:NS_fno3d_pareto_bound}, and yields the best trade-off in terms of accuracy versus computational training time. Surprisingly, it allows us to maintain the accuracy reached by the baseline at the resolution $R = 64$, while decreasing the training time by a factor of $11$ (resp.~$3$) compared to the baseline trained on a $64\times 64$ (resp.~$32\times 32$) grid. We also considered the optimal sample allocation $N_i^* = 2^{2i}N_3$ for $i\in\{1,2\}$ with $\delta^*=2$, $N_3^* = 45$ at resolution $32\times 32$, and plotted as a bordeaux dot in \cref{fig:NS_fno3d_pareto_bound}. Finally, we do not observe a significant difference in terms of accuracy when considering nested sub-sampling over random sampling. Now considering \cref{fig:NS_fno3d_plot_best_worst} (left), we compare ground truth solutions with the approximations obtained by MLMC training algorithm with $m=4$ (resp.~$m=2$) levels and $\delta=4$ (resp.~$\delta=2$) at the time $T=40$. \cref{fig:NS_fno3d_plot_best_worst} displays the evolution of the coarse and pairwise difference training losses composing the telescopic sum in \cref{eq_MLMC} before evaluating the gradients for $m=4$ levels and $\delta = 2$. As expected, the coarse loss is dominating the telescopic sum throughout the training, while the pairwise difference losses across levels decreases with the level. We note a very small contribution from the last level, explaining the improved performance of the magenta curve settings (maximum resolution of $32\times 32$) in \cref{fig:NS_fno3d_pareto_bound}.

\subsection{Message passing neural PDE solver}
In this experiment, we consider the message-passing neural PDE solver architecture introduced by~\cite{brandstettermessage}. This is a type of graph neural network~\cite{kipf2016semi,battaglia2018relational} written in the encoder-processor-decoder framework following earlier works~\cite{battaglia2018relational,sanchez2020learning}. We consider the problem of learning the time-forward velocity map $u(\mathbf{x},t)\mapsto u(\mathbf{x},t+\Delta t)$ associated with the Navier-Stokes equations on a classical fluid flow past a cylinder problem, near the Hopf bifurcation at Reynolds number $\textrm{Re}=200$, which features von Kármán vortex street patterns. This is a well-known benchmark problem for the evaluation of numerical algorithms for incompressible time-dependent Navier-Stokes equations~\cite{schafer1996benchmark}. The equations read as
\begin{equation}
  \begin{aligned}
    \frac{\partial \mathbf{u}}{\partial t} +  (\mathbf{u}\cdot \nabla) \mathbf{u}                                                        - \nabla \cdot \frac{2}{\textnormal{Re}}\boldsymbol{\epsilon}(\mathbf{u}) + \nabla p & = 0,                        &        & \text{in } \Omega \times (0,T], \\
    \nabla \cdot \mathbf{u}                                                                                                                                                                                                   & = 0,                        & \qquad & \text{in } \Omega \times (0,T], \\
    \mathbf{u}(\mathbf{x}, 0)                                                                                                                                                                                                 & = \mathbf{u}_0(\mathbf{x}), &        & \mathbf{x} \in \Omega,          \\
  \end{aligned}
  \label{eq:flow}
\end{equation}
where $\Omega \subset \mathbb{R}^2$, $T=50$, $p: \Omega \times [0, T) \rightarrow \mathbb{R}$ is the pressure field, the Reynolds number is Re = 200, and $\boldsymbol{\epsilon}(\mathbf{u})= (\nabla \mathbf{u} + \nabla \mathbf{u}^\top)/2$. The boundary conditions are defined as follows: $\mathbf{u} = (1, 0)^\top$ on the upper and lower side of $\Omega$, a homogeneous Dirichlet boundary condition on the disk obstacle, and a ``do-nothing" condition on the right side of the domain. We consider fluid fields including laminar flow as well as vortex structures.
The dataset is formed of $N=200$ samples obtained by considering different random inflow conditions on the left boundary, which are split as $150$ for training and the remaining $50$ for testing. The equations are discretized in space using standard Taylor-Hood finite elements for velocity and pressure, which is a stable element pair for the Stokes equation \cite{TAYLOR197373}, and in time by a Crank--Nicolson time-stepping scheme with a timestep of $0.25$. The velocity and pressure solutions are sampled every $\Delta t= 1$ timesteps, and we aim to learn the time-forward operator $\mathbf{u}(\cdot,t)\mapsto \mathbf{u}(\cdot,t+\Delta t)$, by mapping the first $30$ time steps to the next $20$. The finest mesh $M_4$ on which the simulations are performed contains $14631$ triangle elements with $7217$ nodes, and a target element size of $0.03$ near the obstacle and $0.12$ on the boundary of the domain. The computed solutions are projected onto a space of linear finite elements and the values at the mesh nodes are saved. We also generated coarser meshes using the Gmsh meshing software~\cite{geuzaine2009gmsh} with approximately twice fewer nodes from level $i+1$ to level $i$ (specifications are available in \cref{tab:ns_mesh}) to run the MLMC algorithm, and performed numerical simulations on these coarser meshes to acquire the corresponding data. This contrasts with previous experiments of \cref{sec_fno} (and what is usually done in operator learning), where the numerical solutions are acquired on the finest resolution and down-sampled to coarser ones. Here, we aim to evaluate the performance of MLMC training in the challenging settings where the coarse data quality is low (since they are generated by a numerical solver on a coarser spatial discretization).

\begin{table}[htbp]
  \centering
  \caption{Characteristics of the different meshes used in the fluid flow past a cylinder example.}
  \par\vspace{1em}\par 
  \begin{tabular}{lcccc} \toprule[\thick pt] Mesh name & $\#$ nodes & $\#$ elements & mesh size [obstacle] & mesh size [boundary]
               \\ \midrule[\thick pt]
               $M_4$                                 & 7217       & 14631         & 0.05                 & 0.18                 \\
               $M_3$                                 & 3808       & 7760          & 0.07                 & 0.25                 \\
               $M_2$                                 & 1645       & 3387          & 0.10                 & 0.4                  \\
               $M_1$                                 & 806        & 1682          & 0.14                 & 0.6                  \\
               \bottomrule[\thick pt]
  \end{tabular}
  \label{tab:ns_mesh}
\end{table}

We report in \cref{tab:fpc_mppde_baseline} the mean square error (MSE) of the testing loss on the finest mesh $M_4$ after training the MP-PDE graph neural networks on the different mesh resolutions. As expected, the testing error decreases as the training data resolution increases, but at the cost of increasing the computational training time. The last two columns of \cref{tab:fpc_mppde_baseline} report the test loss and average epoch training time (over 200 epochs) when using MLMC with $m=4$ levels and $m=2$ (with resolution $M_4$ and $M_1$ only),  respectively, both with $\delta = 2$. We observe that the two MLMC training strategies yield the best trade-off in terms of accuracy versus computational time (measured by the product of the test loss and timings), outperforming the Pareto curve defined by the baseline performance.

\begin{table}[htbp]
  \centering
  \caption{\textit{Fluid flow past cylinder problem.} Baselines average training time per epoch and test loss when training the graph neural network for varying resolutions $M_i$, along with the results using MLMC training. Best trade-off, defined as training time multiplied by test MSE, is in \textbf{bold}, while second best is \underline{underlined}.}
  \par\vspace{1em}\par 
  \begin{tabular}{lcccccc}
    \toprule[\thick pt]
    Training resolution              & $M_1$ & $M_2$           & $M_3$        & $M_4$ & MLMC (4,2) & MLMC (2,2) \\ \midrule[\thick pt]
    Training epoch time $[s]$        & 0.36  & 0.67            & 1.50
                                     & 1.84  & 0.66            & 0.85                                           \\
    MSE test loss [$\times 10^{-2}$] & 5.0   & 5.3             & 5.7
                                     & 0.5   & \underline{1.1} & \textbf{0.8}                                   \\
    \bottomrule[\thick pt]
  \end{tabular}
  \label{tab:fpc_mppde_baseline}
\end{table}

\subsection{Mesh-free convolutional neural network}
\label{sec_mfconv}
The architecture considered is a resolution-invariant CNN specifically designed for mesh-based inputs, introduced in \cite{zakariaei2025multiscale, ruthotto2020deep}. The main idea is to define convolution operations without relying on a mesh of the domain $\Omega \subset \mathbb{R}^2$, adopting a mesh-free approach instead. More precisely, the mesh-free convolution reads as
\begin{equation}
  \label{eq:mfconv}
  C(\boldsymbol{\xi}) * a(\mathbf{x}) = (\gamma_0 + [\gamma_x, \gamma_y]^T \nabla)\tilde{v}(\tau, \mathbf{x}),
\end{equation}
where $\tilde{v}(\tau, \mathbf{x})$ is obtained by the solution of the following parabolic PDE
\begin{equation}
  \label{eq:mfconv_pde}
  \begin{cases}
    \displaystyle \frac{\partial \tilde{v}(t, \mathbf{x})}{\partial t} = \mathcal{L}(\boldsymbol{\alpha})\tilde{v}(t, \mathbf{x}) \quad t \in (0, \tau], \\
    \tilde{v}(0, \mathbf{x}) = \mathcal{R}(\boldsymbol{\theta})a(\mathbf{x}).
  \end{cases}
\end{equation}
The elliptic differential operator in \cref{eq:mfconv_pde} is defined as
\begin{equation*}
  \mathcal{L}(\boldsymbol{\alpha}) \;=\; \alpha_{xx} \,\frac{\partial^2 v}{\partial x^2}
  \;+\; 2\,\alpha_{xy}\,\frac{\partial^2 v}{\partial x\,\partial y}
  \;+\;\alpha_{yy}\,\frac{\partial^2 v}{\partial y^2},
\end{equation*}
where $\alpha_{xx}, \alpha_{yy} > 0$ and $\alpha_{xx} \alpha_{yy} - \alpha_{xy}^2 > 0$ thus ensuring that $\mathcal{L}$ is elliptic and invertible. The vector
$\boldsymbol{\xi} =[\tau, \boldsymbol{\alpha}, \boldsymbol{\gamma}] \in \mathbb{R}^7$ collects all the trainable parameters. The input field $a$ is mapped to the embedding $\tilde{v}$ by means of the operator $\mathcal{R}: \mathcal{A} \rightarrow \mathbb{R}^{d_v}$ consisting of a convolution operation \cref{eq:mfconv} with $1 \times 1$ filters. In this way, the initial condition of \cref{eq:mfconv_pde} is obtained and the parabolic PDE is solved. Finally, the output of the convolution is obtained by taking the derivative of $\tilde{v}(\tau, \mathbf{x})$ in the direction $\gamma$. So, introducing a suitable uniform discretization of the domain $\Omega$, solving \cref{eq:mfconv_pde} by means of Fast Fourier Transform (FFT) and differentiating the solution in the direction $\boldsymbol{\gamma}$ actually consists of applying the mesh-free convolution \cite{zakariaei2025multiscale}.

\paragraph{Advection-diffusion equation}
We consider an advection-diffusion equation on a two-dimensional rectangle $\Omega = [-5, 5] \times [-4, 4]$:
\begin{equation}
  \label{eq:AD}
  \begin{aligned}
    - \nabla \cdot (\nu(\mathbf{x}) \nabla u(\mathbf{x})) + \mathbf{b} \cdot \nabla u(\mathbf{x}) & = f(\mathbf{x}), &  & \mathbf{x}\in\Omega,     \\
    u(\mathbf{x})                                                                                 & = 0,             &  & \mathbf{x} \in \Gamma_D, \\
    \displaystyle \nu \frac{\partial u}{\partial n}                                               & = 0,             &  & \mathbf{x} \in \Gamma_N.
  \end{aligned}
\end{equation}
This equation models the transport and diffusion of the pollutant released by industrial chimneys into a plane \cite{M2AN_2005__39_5_1019_0}. Here, $\mathbf{b} = (0.0025, 0.0025)^\top$, $f(\mathbf{x})= \sum_{i = 1}^4 f_i \chi_{D_i}(\mathbf{x})$, where $f_i \geq 0$, and $D_i$ the spatial region representing the $i$-th chimney and $f_i$ its emission rate (normalized with respect to an appropriate emission volume). We impose a homogeneous Dirichlet condition over the inflow boundary $\Gamma_D = \{ \mathbf{x} \in \partial \Omega: \mathbf{b} \cdot \mathbf{n}(\mathbf{x}) < 0 \}$, where $\mathbf{n}(\mathbf{x})$ is the unit vector directed outward, and homogeneous Neumann condition on the outflow boundary $\Gamma_N = \partial \Omega \setminus \Gamma_D$. For the expression of the diffusion field $\nu$ we refer to \cite{M2AN_2005__39_5_1019_0}. \cref{eq:AD} is discretized in space using linear finite elements by considering $\{ R_i \times R_i \}_{i = 1}^3 = \{75 \times 75, 150 \times 150, 300 \times 300 \}$ space points. The dataset consists of $N = 1024$ samples obtained by drawing instances of $x_c^i \sim \mathcal{U}(-3, -2)$ and $r^i \sim \mathcal{U}(0.1, 0.3)$, parametrizing the regions $D_i$, for $i\in\{1, \ldots, 4\}$.

We validate the proposed algorithm by employing mesh-free convolutional neural networks on the test cases \cref{eq:darcy,eq:AD}. In particular, to take into account advection and incorporate the knowledge we have about the system in \cref{eq:AD}, we add to the mesh-free convolutional kernel a term of the form \cite{ruthotto2020deep}:
\begin{equation*}
  \psi_x[-1, 0, 1] \quad \textnormal{and} \quad \psi_y[1, 0, -1],
\end{equation*}
that is we add a second convolutional layer by fixing their weights equal to the second order finite difference discretization of the advection term, and we multiply the output by a trainable parameter in each direction of the domain. Then, the vector of all trainable parameters becomes $\boldsymbol{\xi} =[\tau, \boldsymbol{\alpha}, \boldsymbol{\gamma}, \boldsymbol{\psi}] \in \mathbb{R}^9$ for each mesh-free convolutional layer.

\begin{table}[htbp]
  \centering
  \caption{\textit{Darcy problem.} (a) Baselines training time per epoch and test loss by varying the resolution $R$. (b) MLMC training time per epoch and test loss by varying $\delta$ and $m=3$.}
  \par\vspace{1em}\par 
  \begin{tabular}{lccccc}
    \toprule[\thick pt]
    Resolution             & $R = 15$ & $R = 30$ & $R = 60$ & $R = 120$ & $R = 241$ \\ \midrule[\thick pt]
    Train epoch time $[s]$ & 0.81     & 0.83     & 0.91     & 1.26      & 3.96      \\         Test loss  [$\times 10^{-2}$]             & 15.2  & 2.5  & 2.0  & 2.0   & 1.8   \\ \bottomrule[\thick pt]
  \end{tabular}
  \put(-320,15){(a)}
  \par\vspace{1em}\par 
  \begin{tabular}{lcccccc}
    \toprule[\thick pt]
    Parameter $\delta$           & $\delta = 8$  & $\delta = 4$  & $\delta = 2$ & $\delta = 1.5$ & $\delta = 1$ & $\delta^* = 2$ \\ \midrule[\thick pt]
    Train epoch time $[s]$       & \textbf{0.19} & \textbf{0.55} & 1.25         & 1.66           & 2.48         & \textbf{0.60}  \\
    Test loss [$\times 10^{-2}$] & \textbf{7.8}  & \textbf{4.7}  & 3.3          & 3.2            & 3.1          & \textbf{2.5}   \\ \bottomrule[\thick pt]
  \end{tabular}
  \put(-320,15){(b)}
  \label{tab:darcy_para_baseline}
\end{table}

\begin{figure}[htbp]
  \centering
  \includegraphics[scale = 0.3]{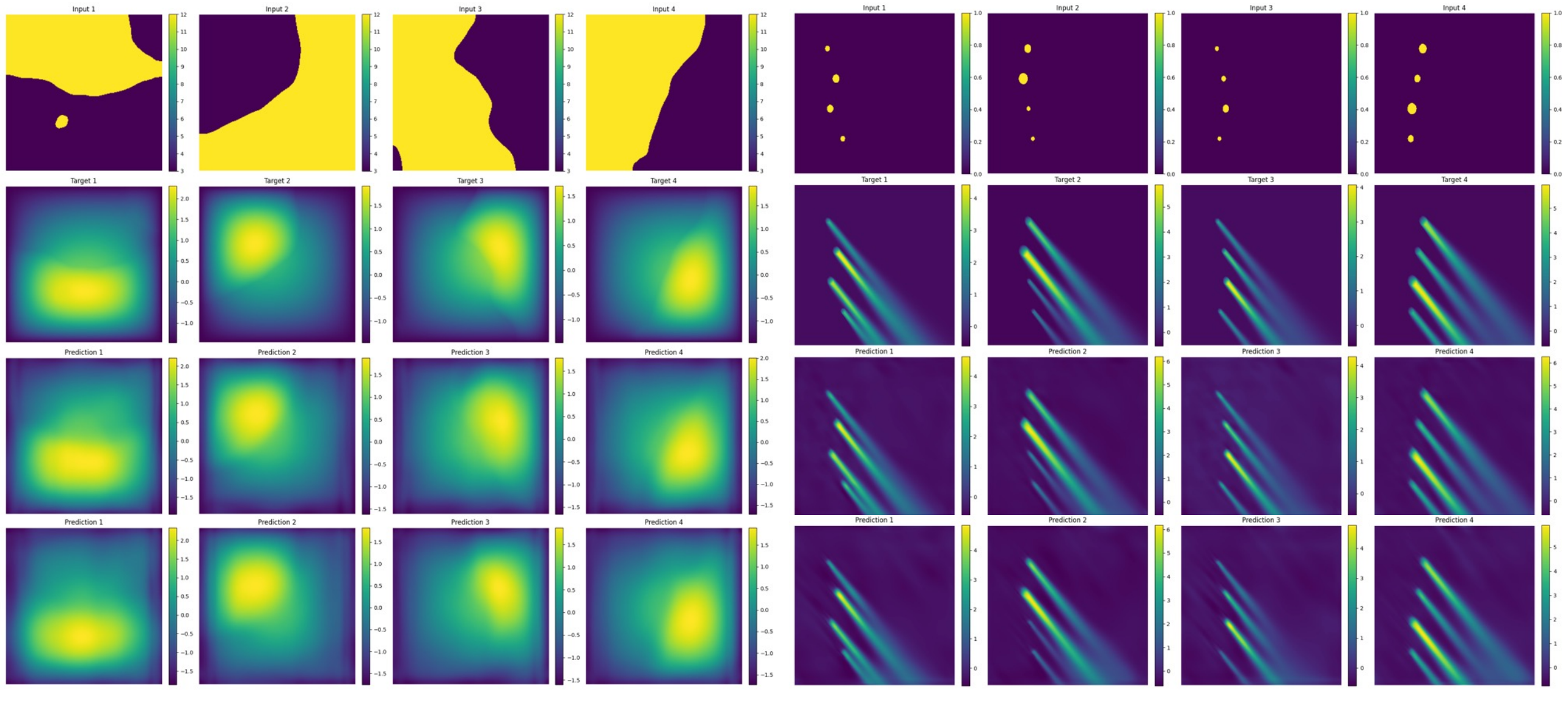}
  \caption{\textit{Darcy (left) and advection-diffusion (right) problems.} Comparison between the ground truth data (second row), the SGD solution (third row) and the MLMC approximation (fourth row) for different source inputs (first row).}
  \label{fig:AD_para_comparison_field}
\end{figure}

\paragraph{Test on Darcy problem} To obtain the baselines, we first run standard SGD to approximate the solution operator on the resolutions $\{R_m\}_{m = 1}^5 = \{ 241, 120, 60, 30, 15 \}$ and the results are the ones reported in \cref{tab:darcy_para_baseline}(a). The training time per epoch and the testing loss values obtained by means of our proposed training paradigm, with $m=3$ levels and hierarchy as pairing strategy, are gathered in \cref{tab:darcy_para_baseline}(b). As expected, by increasing the multiplier $\delta$ the test loss increases as the epoch train time decreases because a smaller amount of fine training data is used. We also report the test related to the use of the optimal number of samples per resolution $N_{i}^*$ (\cref{prop:MLMC} with $k=1$ and $d=2$). Moreover, we select $\delta = \delta^* = 2$. This configuration is the one providing the best performance, in terms of efficiency, with respect to the baseline with $R = 30$ (the baseline resolution providing a comparable accuracy), and it is related to the fact that even if $N_i^* \approx 2^{-2i}N_{i-1}^*$ we are employing $\delta = 2$ instead of $\delta = 8$; indeed, in the latter case, during one epoch less batches of fine resolution data are seen with respect to the number of batches of coarse data. As the value for $\delta$ increases the gap between the number of batches of fine and coarse data increases making the gradients' estimates more and more biased towards large-scales, or in the same way, the exposure to fine-scale corrections is limited. In \cref{fig:AD_para_comparison_field}, we report the comparison among the ground truth data and the approximation obtained by means of traditional SGD, both with $R = 241$, and the MLMC approximated solution, with hierarchy pairing strategy, $m^*=3$ and $\delta^* = 2$, for different testing instances of the diffusion coefficients.

\paragraph{Test on advection-diffusion} To obtain the baseline quantities for the advection-diffusion test-case \cref{eq:AD}, we just consider resolution $R = 300$, since smaller ones did not provide an acceptable accuracy, considering the current model and expressivity. So, in this case, the goal is to show that coarser data can be actually exploited in the MLMC paradigm to capture large-scale effects together with finer correction terms. The baseline epoch training time is 8.97$s$ while the testing loss is equal to 0.08$\%$. In \cref{fig:AD_para_comparison_field}, we compare the ground truth data (second row), the baseline SGD approximation (third row), and the approximated solution obtained by means of MLMC with $m = 3$ and $\delta = 2$, resulting in a test loss of 2.1$\%$ and a computational speed-up equal to 4, for different instances of the source term.

\subsection{Geometry-informed neural operator transformer} \label{sec_ginot}
\begin{figure}[htbp]
  \centering
  \begin{minipage}{0.48\textwidth}
    \centering
    \begin{overpic}[width=1.\textwidth]{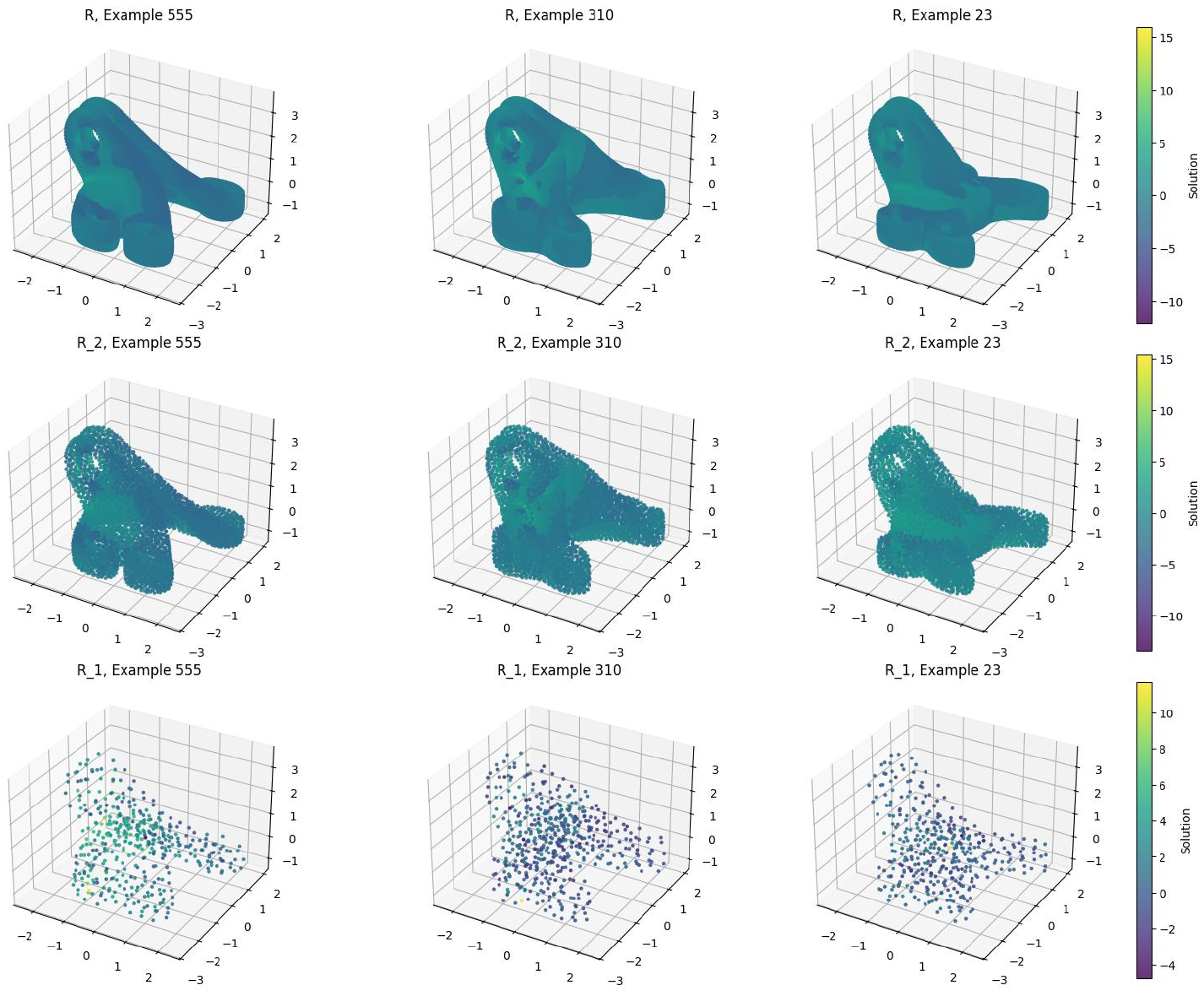}
      \put(0,65){(a)}
      \put(100,65){(b)}
    \end{overpic}
  \end{minipage}
  \hspace{0.01\textwidth}
  \begin{minipage}{0.48\textwidth}
    \centering
    \begin{overpic}[width=1.\textwidth]{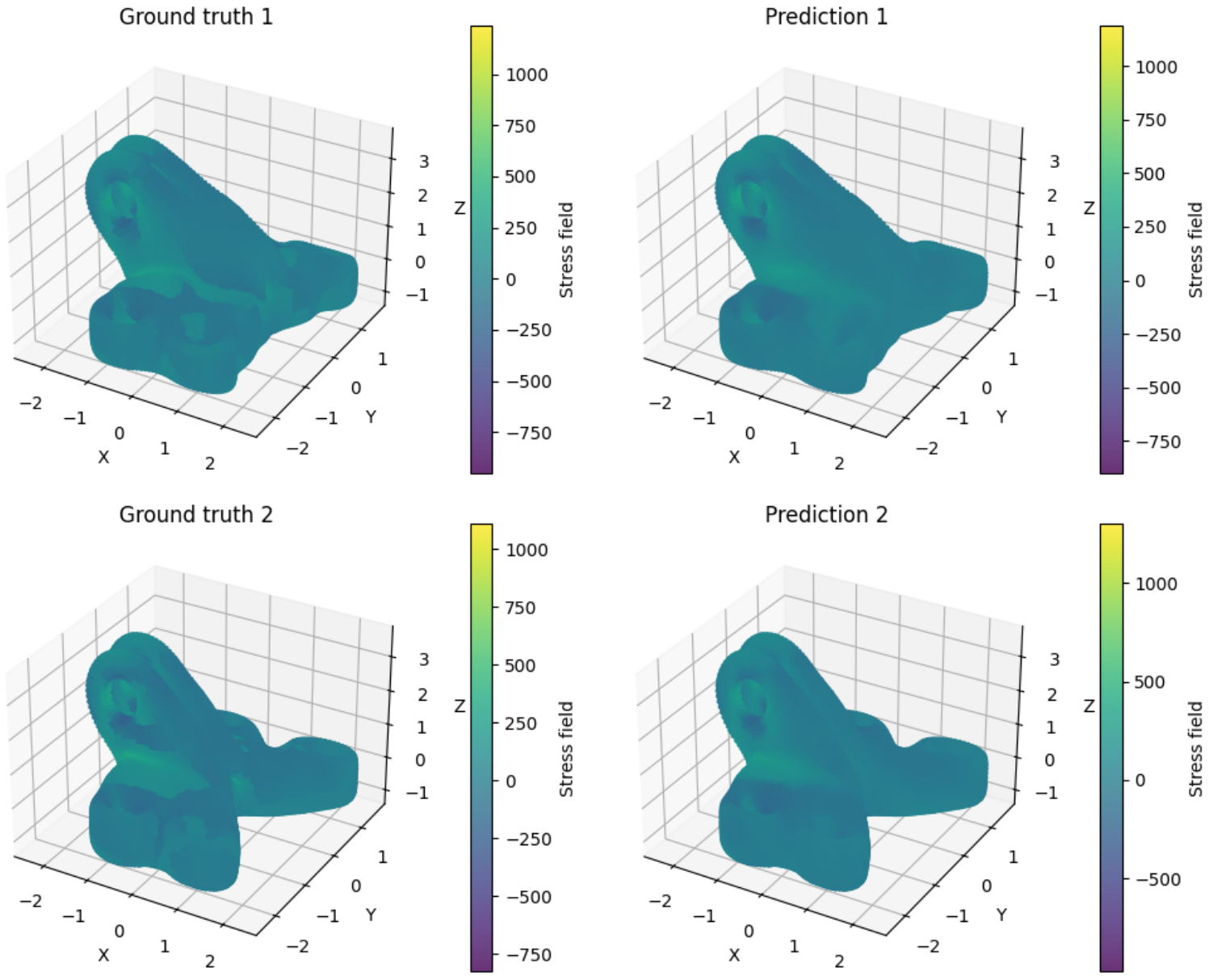}
    \end{overpic}
  \end{minipage}
  \caption{\textit{DeepJEB test problem.} (a) Three sample geometries from the DeepJEB dataset at resolution $R_{\text{test}}$, $R_2$, and $R_1$. (b) Comparison between the ground truth data (first column), and the MLMC solutions (second column) for different shapes.}
  \label{fig:JEB_GINOT_comparison_field}
\end{figure}

GINOT ~\cite{liu2025geometryinformedneuraloperatortransformer} is a neural operator architecture based on transformer networks~\cite{vaswani2017attention} that is designed to learn mappings between complex geometries (represented as CAD meshes or cloud of points) to corresponding solution fields (see \cref{fig:JEB_GINOT_comparison_field}). It consists of two main components: a geometry encoder and a solution decoder, which are combined by the cross-attention mechanism of the transformer network.

In this numerical experiment, we apply the MLMC neural operator training algorithm to the GINOT architecture on the Deep jet engine bracket (DeepJEB) dataset~\cite{10.1115/1.4067089} containing 3D large-scale jet engine brackets and corresponding von Mises stress fields under vertical loading. The DeepJEB datatest is publicly accessible and comprises $N = 2138$ samples of complex 3D jet engine brackets with arbitrary shapes and sizes. More precisely, the dataset was created using deep generative models to perform data augmentation on the simulated jet engine bracket (SimJEB) dataset~\cite{cgf.14353} thus enhancing the diversity of shapes and the performance features by means of AI. SimJEB is one of the largest publicly available dataset for jet engine bracket, and includes 381 hand-designed CAD models, along with finite element simulation results of von Mises stress and displacement fields under different loading conditions. Following \cite{liu2025geometryinformedneuraloperatortransformer}, we focus on the approximation of the von Mises stress data under vertical loading.
The DeepJEB dataset volume mesh nodes ranges from $18085$ to $55009$, which we refer to as testing resolution $R_{\text{test}}$ in this paper. We use it to evaluate the super-resolution capabilities of the trained transformer model~\cite{liu2025geometryinformedneuraloperatortransformer}.
To apply the MLMC training algorithm, we generate a similar dataset at a coarse level of resolution $R_1 \in [282,859]$ and a fine level of resolution $R_2 \in [4521,13752]$ by downsampling the original dataset. The dataset downsampling was performed using the PointNet++~\cite{qi2017pointnet} algorithm exploiting the \texttt{pool.fps} command from the \texttt{torch\_geometric} library~\cite{fey2019fast} with a default sampling ratio of $0.5$.

\begin{table}[htbp]
  \centering
  \caption{Testing loss achieved by training GINOT on the coarse, fine, and test resolutions along with the corresponding average training time per epoch (over $500$ epochs). The last row reports the performance of the MLMC training algorithm with $m=2$ levels of resolution $R_1$ and $R_2$ and a batch multiplier $\delta = 2$.}
  \par\vspace{1em}\par 
  \begin{tabular}{lcc}
    \toprule[\thick pt]
    Training method                   & Test loss on $R_{\text{test}}$ & Training epoch time $[s]$ \\
    \midrule[\thick pt]
    Coarse resolution $R_1$           & 33.4\%                         & 16.9                      \\
    Fine resolution $R_2$             & 22.7\%                         & 40.2                      \\
    Test resolution $R_{\text{test}}$ & 18.6\%                         & 101.2                     \\
    \midrule[\thick pt]
    MLMC on $R_1+R_2$                 & 26.3\%                         & 20.1                      \\
    \bottomrule[\thick pt]
  \end{tabular}
  \label{tab:baseline}
\end{table}

We consider the same hyperparameters as in \cite{liu2025geometryinformedneuraloperatortransformer} for the GINOT architecture, except for the number of attention heads in the encoder and decoder, which we set to $4$ instead of $1$, encoder/decoder cross-attention layers (set to $4$) and encoder self-attention layers (set to $4$). By using a larger architecture, we are able to improve the performance of the GINOT model on the test resolution and reach a normalized mean square error of $18.6\%$, which is much lower than $35.6\%$ reported in \cite{liu2025geometryinformedneuraloperatortransformer}. Sample geometries from the dataset at different resolutions are depicted in \cref{fig:JEB_GINOT_comparison_field}(a). We train the GINOT on the DeepJEB dataset at different resolutions $R_1$, $R_2$, and $R_{\text{test}}$ over $500$ epochs and report the average training time per epoch along with the test loss on the resolution $R_{\text{test}}$ in \cref{tab:baseline}. We note that we slightly modified the GINOT architecture and were able to outperform the testing error of $35.6\%$ on the resolution $R_{\text{test}}$ reported by \cite{liu2025geometryinformedneuraloperatortransformer}.

\begin{figure}[htbp]
  \centering
  \includegraphics[width=\textwidth]{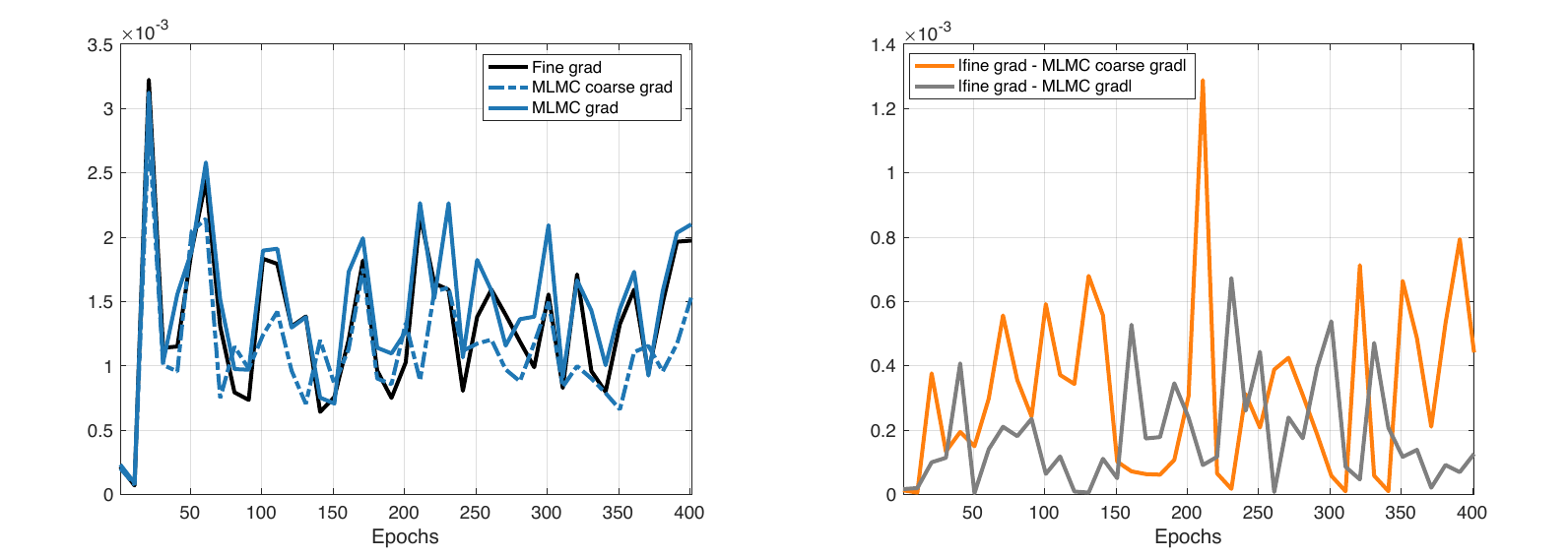}
  \vspace{-.05\textwidth}
  \caption{\textit{DeepJEB test problem.} Left: Comparison between the norm of the fine gradients, the norm of the MLMC coarse gradients and MLMC gradients approximation over the same batch. Right: Absolute error between the fine gradients, and the gradients of the coarse and telescopic sum given by \cref{eq:MLMC_batch} over the same batch.}
  \label{fig:JEB_GINOT_gradient_analysis}
\end{figure}

In the last row of \cref{tab:baseline}, we show the performance (on the finest resolution $R_{\text{test}}$) of the MLMC training algorithm with the two levels of resolution $R_1$ and $R_2$, together with with the corresponding computational training time per epoch. In this complex task, the proposed MLMC approach achieves a testing loss of $26.3\%$ with a training time of $20.1$ seconds per epoch, which improves upon the Pareto frontier of the training time/accuracy trade-off, between the coarse ($R_1$) and fine ($R_2$) resolution training baselines. We note that the training time is nearly halved compared to the fine resolution training on $R_2$ while losing only the 15$\%$ in terms of accuracy. In \cref{fig:JEB_GINOT_comparison_field}(b), we display one ground truth sample at resolution $R_{\text{test}}$ and its MLMC approximation obtained by considering two levels of resolutions $m=2$, with a nested sub-sampling as pairing strategy, $1723$ and $213$ samples at level one and two (optimal sample progression), and a batch multiplier of $\delta = 2$. Finally, in \cref{fig:JEB_GINOT_gradient_analysis}, we display the corresponding approximation of the fine gradients as we include the fine resolution correction term in the telescopic sum of \cref{eq:MLMC_batch}.

\section{Conclusions}
Neural operators have shown remarkable promise in the approximation of operators associated with PDEs, but their practical adoption is hindered by the computational cost of high-resolution training. We introduced a mathematically rigorous MLMC approach to efficiently train neural operators by approximating fine-resolution gradients through a telescopic sum formulation across mesh hierarchies or function discretizations. Our implementation is characterized by a careful design of resolution-aware data structures and optimized memory management. Our numerical results, across a variety of state-of-the-art neural operator architectures and complex benchmarks, demonstrated that the proposed MLMC training algorithm accelerates training while maintaining the same accuracy.

Current limitations, that can be addressed by future developments, are related to the need for sufficiently complex model architecture in order to benefit from the multi-resolution approach. The optimal sampling distribution depends on problem-specific variance decay rates that may need fine-tuning for each application. Finally, approximation errors from coarse representations might be dominating in problems with extremely sharp features or discontinuities, and potentially reduce the efficiency of the MLMC approach. Additional future developments could include dynamically adjusting the sampling distributions across resolutions during training. Finally, MLMC is a flexible approach, which leads itself to pretraining and finetuning applications, and possibly extensions to online learning scenarios, thus enabling efficient adaptation of pretrained models to new data distributions with minimal computational overhead.

\section*{Acknowledgements}
This work originates from an Oberwolfach workshop on Deep Learning for PDE-based Inverse Problems. SF is member of the Gruppo Nazionale Calcolo Scientifico-Istituto Nazionale di Alta Matematica (GNCS-INdAM) and acknowledges the project ``Dipartimento di Eccellenza'' 2023–2027, funded by MUR. SF acknowledges the Istituto Nazionale di Alta Matematica (INdAM) for the financial support received (Scholarships for visiting periods abroad, 2023–2024) and the project FAIR (Future Artificial Intelligence Research), funded by the NextGenerationEU program within the PNRR-PE-AI scheme (M4C2, Investment 1.3, Line on Artificial Intelligence). CBS and SF acknowledge the EPSRC program grant in ``The Mathematics of Deep Learning'', under the project EP/V026259/1. NB was supported by the Office of Naval Research (ONR), under grant N00014-23-1-2729.

\bibliographystyle{elsarticle-num}
\bibliography{references}

\end{document}